\documentclass[10pt,letterpaper,journal]{IEEEtran}
\usepackage[T1]{fontenc}
\usepackage[utf8]{inputenc}
\usepackage{amsmath,amssymb,amsfonts}
\usepackage{graphicx,array,booktabs,tabularx}
\usepackage{textcomp,stfloats,cite}
\usepackage{placeins}
\usepackage{url}
\usepackage[hidelinks]{hyperref}
\hypersetup{
  pdftitle={FeaXDrive2: Body--Scene Spatial Grounding for Feasibility-Aware Generative Driving},
  pdfsubject={arXiv preprint}
}
\usepackage{microtype}
\allowdisplaybreaks
\begin{document}
\title{FeaXDrive2: Body--Scene Spatial Grounding for Feasibility-Aware Generative Driving}
\hypersetup{pdfauthor={Bo Leng, Baoyun Wang, Wenxin Wen, Ming Liu, Weiqi Zhang,
Ran Yu, Chongfeng Wei, Zhuoren Li, Lu Xiong}}
\author{Bo~Leng, Baoyun~Wang\textsuperscript{*}, Wenxin~Wen, Ming~Liu, Weiqi~Zhang,\\
Ran~Yu, Chongfeng~Wei, Zhuoren~Li\textsuperscript{*}, and~Lu~Xiong%
\thanks{\textsuperscript{*}\textit{Corresponding authors: Baoyun Wang and Zhuoren Li.}
(e-mail: wangby@tongji.edu.cn; zhuoren@hku.hk).}%
\thanks{Bo Leng, Baoyun Wang, Wenxin Wen, Ming Liu, Weiqi Zhang, Ran Yu,
and Lu Xiong are with the College of Automotive and Energy Engineering,
Tongji University, Shanghai 201804, China
(e-mail: lengbo@tongji.edu.cn;
2532914@tongji.edu.cn; 2110215@tongji.edu.cn; 2511517@tongji.edu.cn;
2433113@tongji.edu.cn; xiong\_lu@tongji.edu.cn).}%
\thanks{Zhuoren Li is with the Department of Civil Engineering,
The University of Hong Kong, Pokfulam Road, Hong Kong, China.}%
\thanks{Chongfeng Wei is with the James Watt School of Engineering,
University of Glasgow, Glasgow G12 8QQ, United Kingdom
(e-mail: Chongfeng.Wei@glasgow.ac.uk).}%
}

\maketitle
\begin{abstract}
Generative driving planners may produce trajectories that appear feasible at their reference points but violate environmental constraints over the full vehicle body. We propose FeaXDrive2, an end-to-end generative planning framework that enhances vehicle-embodied spatial feasibility (VESF) through body--scene spatial grounding. The proposed method constructs spatial queries from actual vehicle-body corners and dynamically updates visual sampling locations using evolving trajectory estimates across Flow steps. Retrieved local visual features condition subsequent Action Flow-DiT velocity predictions, allowing trajectory generation to guide spatial information retrieval. Rollout-Consistent Safety Fine-Tuning (RC-F) combines high-risk trajectory supervision with reference behavior preservation, followed by planning-supervised VLM--Action Alignment. Vehicle-geometry counterfactual analysis reveals safety differences under identical trajectories but varying vehicle geometry. On NAVSIM, FeaXDrive2 reduces at-fault collisions by 36.5\% relative to the base planner and improves PDMS from 87.0 to 88.8. Under NAVSIM v2 evaluation, EPDMS increases from 86.5 to 88.3. Ablation and body-clearance analyses further demonstrate the safety benefits of body--scene spatial grounding.
\end{abstract}

\section{Introduction}
\label{sec:I}

The safety of end-to-end driving planning depends not only on the motion feasibility of planned trajectories but also on whether the full vehicle-body occupancy along these trajectories satisfies environmental constraints. Recent advances in generative models and vision-language-action (VLA) models have improved end-to-end driving planning. DriveVLM \cite{tian2025drivevlm} and EMMA \cite{Hwang2025EMMA} incorporate vision-language modeling into driving through hierarchical planning and multimodal task modeling, respectively. DiffusionDrive \cite{liao2025diffusiondrive} employs diffusion models to generate multimodal driving trajectories, while Alpamayo-R1 \cite{wang2025alpamayo} connects driving reasoning with continuous action prediction. Nevertheless, even semantically reasonable and smooth trajectories may result in vehicle-body contact during narrow-road passing, close-range obstacle avoidance, or driving near road boundaries. We define vehicle-embodied spatial feasibility (VESF) as the extent to which the full vehicle-body occupancy induced by a planned trajectory satisfies environmental spatial constraints under given vehicle geometry and traffic conditions. VESF thus depends jointly on vehicle geometry, trajectory poses, and environmental constraints.

\begin{figure*}[!t]
\centering
\includegraphics[width=0.95\textwidth]{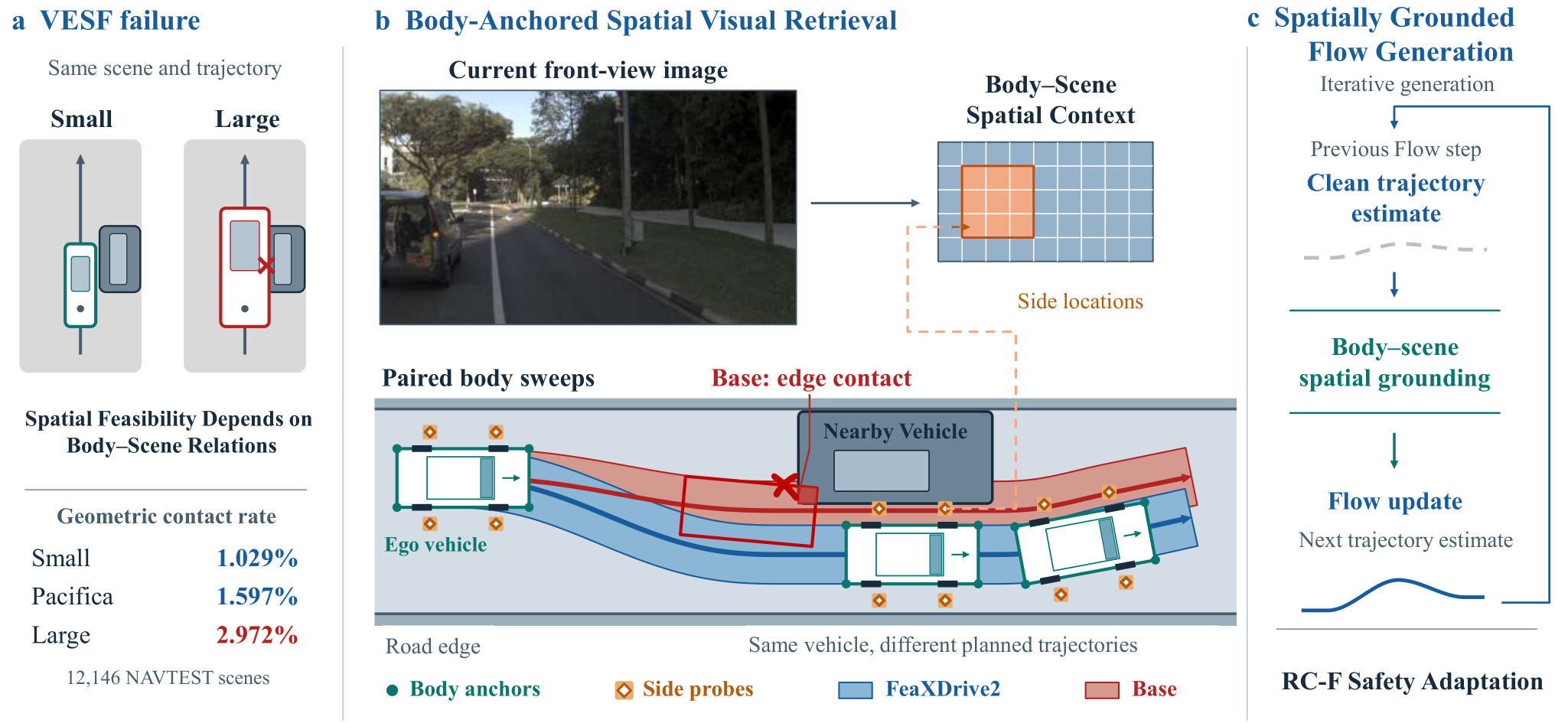}
\caption{Motivation and core idea of FeaXDrive2. (a) Vehicle geometry changes contact outcomes under identical trajectories. (b) Vehicle-body anchors and evolving trajectory estimates guide local visual feature retrieval. (c) Retrieved body--scene context conditions subsequent Flow updates, while RC-F enhances planning safety during training.}
\label{fig:1}
\end{figure*}

Vehicle-geometry counterfactual analysis further quantifies the influence of full-body occupancy on planning safety. On the NAVTEST scenarios of NAVSIM \cite{dauner2024navsim}, we keep the rear-axle trajectories predicted by the base planner and the simulated motion states unchanged, varying only the vehicle geometry used for evaluation. The resulting geometric contact rates for Small, Pacifica, and Large vehicle configurations are 1.029\%, 1.597\%, and 2.972\%, respectively, with Large exhibiting an 86.08\% increase relative to Pacifica (Fig.~\ref{fig:1}(a)). These results demonstrate that vehicle geometry substantially affects geometric contact outcomes under identical scenes and trajectories. Further analysis reveals that changes in vehicle geometry can also lead to safety-metric failures without direct geometric contact, indicating that VESF involves not only geometric contact but also drivable-area constraints and potential collision risks.

A key challenge in VESF modeling is to establish dynamic spatial relationships between vehicle-body occupancy and the surrounding environment and incorporate them into trajectory generation. For a rigid vehicle with known geometry, its body occupancy can be determined from the trajectory reference pose and vehicle dimensions. However, identifying the environmental constraints relevant to this occupancy from visual observations remains challenging. Even within the same traffic scene, different candidate trajectories induce different body occupancies, changing which vehicle regions are subject to spatial constraints and which environmental information is relevant to planning. UniAD \cite{hu2023uniad} and SparseDrive \cite{sun2025sparsedrive} connect perception, prediction, and planning through task queries and sparse scene representations, respectively, but these interactions do not explicitly establish correspondences between candidate-dependent vehicle-body occupancy and local image features. Therefore, visual information selection should account for both vehicle geometry and planned trajectories to identify local regions relevant to safe vehicle passage.

Prior studies have explored trajectory-guided visual retrieval and feasibility constraints in generative planning, but how these mechanisms can account for full vehicle-body occupancy remains underexplored. Think Twice \cite{Jia2023ThinkTwice} retrieves features around predicted future positions to refine driving decisions, while HiP-AD \cite{tang2025hipad} employs trajectory-guided deformable attention to retrieve image features, demonstrating that candidate actions can guide environmental information selection. FeaXDrive \cite{wang2026feaxdrive} improves the kinematic and drivable-area feasibility of generated trajectories through adaptive curvature constraints and drivable-area guidance based on full vehicle-body occupancy. GuideFlow \cite{liu2026guideflow} incorporates constraint guidance and energy-based modeling into Flow Matching. Building on these studies, we investigate how vehicle geometry can define body-anchored spatial queries, while evolving trajectory estimates dynamically determine visual retrieval locations. The resulting body--scene spatial context can then condition subsequent trajectory generation to enhance VESF.

For Flow Matching-based driving planners, body--scene spatial grounding must also be translated into safe planning outcomes through the multi-step generation process. Flow Matching \cite{lipman2023flowmatching} learns a velocity field at sampled intermediate states, while the final trajectory is obtained through numerical integration. When intermediate trajectory estimates are used to update visual retrieval locations, evolving generation states also influence subsequent visual conditioning. Consequently, reducing local prediction errors does not necessarily ensure the safety of the final generated trajectory. DRaFT \cite{clark2024draft} investigated the direct optimization of differentiable rewards through diffusion sampling, motivating optimization strategies that account for generated outcomes. Meanwhile, R2SE \cite{liu2026r2se} investigated targeted adaptation to challenging driving scenarios while preserving previously acquired driving capabilities. These considerations motivate safety optimization that accounts for actual generated trajectories while improving high-risk planning behaviors and preserving established safe behaviors.

To address these challenges, we propose FeaXDrive2, an end-to-end generative driving planning framework designed to enhance VESF (Fig.~\ref{fig:1}). The framework employs a pretrained vision-language model for multimodal scene conditioning and learns a base trajectory generator through Flow Matching with a DiT backbone \cite{peebles2023dit}. Safety evaluation of the base planner identifies high-risk scenarios, while vehicle-geometry counterfactual analysis and body clearance measurements further identify challenging safe scenarios, yielding stratified training samples. FeaXDrive2 introduces body-anchored, trajectory-conditioned spatial visual retrieval to establish dynamic body--scene spatial grounding from actual vehicle geometry and evolving trajectory estimates. Rollout-Consistent Safety Fine-Tuning (RC-F) combines high-risk trajectory supervision with reference behavior preservation throughout the multi-step generation process to enhance planning safety. VLM--Action Alignment then jointly optimizes language-model LoRA parameters and selected Action Flow-DiT parameters through planning supervision, aligning multimodal representations with trajectory generation.

Extensive experiments on NAVSIM validate the planning performance and safety improvements of FeaXDrive2. On NAVSIM v1 NAVTEST, FeaXDrive2 reduces at-fault collision scenarios from 181 to 115, a 36.5\% reduction, and improves PDMS from 87.0 to 88.8 under single-candidate inference. On the same scenarios evaluated using the NAVSIM v2 single-stage protocol \cite{cao2025pseudosimulation}, EPDMS increases from 86.5 to 88.3. Component ablations and vehicle-body clearance analysis further examine the contributions of spatial grounding and safety optimization.

The main contributions of this work are summarized as follows:

\begin{enumerate}
\item We propose FeaXDrive2, an end-to-end generative driving planning framework for enhancing VESF. The framework uses vehicle-body occupancy to guide both environmental information acquisition during trajectory generation and the construction of safety supervision, integrating body--scene spatial grounding with safety-oriented learning of the generative planner.

\item We develop a body--scene spatial grounding mechanism and RC-F. Vehicle-body geometry defines query anchors, while evolving trajectory estimates update the corresponding visual retrieval locations to acquire body-relevant environmental information. RC-F further combines high-risk trajectory supervision with reference behavior preservation to improve planning safety and mitigate degradation of established safe behaviors.

\item We characterize and quantify VESF-related failures in end-to-end driving planning. Controlled vehicle-geometry counterfactual analysis quantifies geometric contacts and non-contact safety-metric failures under identical scenes and rear-axle trajectories. Planning evaluations, ablation studies, and vehicle-body clearance analysis further validate the effectiveness of FeaXDrive2.

\end{enumerate}

\section{Related Work}
\label{sec:II}

To contextualize VESF enhancement, this section reviews three related research directions: end-to-end generative driving planning, spatial grounding and physical feasibility modeling, and safety fine-tuning of pretrained planners. These directions provide the foundations for trajectory generation, spatial information interaction, and safety optimization.

\subsection{End-to-End Generative Driving Planning}
\label{sec:II-A}

Multimodal scene representation and action learning provide the foundation for end-to-end driving planning. Early studies established mappings from observations to driving behaviors through conditional imitation learning \cite{Codevilla2018CIL}. Subsequently, TransFuser employed attention mechanisms to fuse multisensor features \cite{Chitta2023TransFuser}, while UniAD \cite{hu2023uniad} and SparseDrive \cite{sun2025sparsedrive} connected perception, motion prediction, and planning through structured scene representations, enabling closer interaction between scene information and driving actions. IPP* coordinates predicted and planned trajectories through safety checks and game-theoretic iteration \cite{rong2026ipp}. Vision-language models further introduced semantic understanding and reasoning into driving planning. DriveVLM integrated scene analysis with hierarchical planning \cite{tian2025drivevlm}, EMMA unified multimodal driving inputs and task outputs \cite{Hwang2025EMMA}, and Alpamayo-R1 connected causal driving reasoning with continuous action prediction \cite{wang2025alpamayo}. VLM-Driver also links scene interpretation and behavior reasoning to waypoint prediction \cite{zhao2026vlmdriver}. These approaches enhanced the use of environmental observations for planning through structured scene representations and semantic conditioning.

Diffusion models \cite{ho2020ddpm,peebles2023dit} and Flow-based generators \cite{lipman2023flowmatching,liu2023rectifiedflow} provide foundations for iterative generative modeling. In autonomous driving, DiffusionDrive improved trajectory generation efficiency through truncated diffusion \cite{liao2025diffusiondrive}, while Diffusion Planner jointly generated future trajectories for the ego vehicle and surrounding traffic participants with support for constraint guidance \cite{Zheng2025DiffusionPlanner}.

Overall, these studies establish a foundation for continuous trajectory generation conditioned on multimodal inputs, motivating further investigation into how candidate trajectories can guide the retrieval of spatially relevant scene information.

\subsection{Spatial Grounding and Physical Feasibility in Driving}
\label{sec:II-B}

Geometric references connect visual features to physical locations. Deformable DETR samples sparse features around reference points \cite{zhu2021deformable}, while DETR3D projects 3D reference points onto multiview image features using camera calibration \cite{wang2022detr3d}. These operations provide a basis for spatial visual retrieval, although their original objectives concern perception rather than vehicle-body feasibility during motion.

Trajectory-conditioned retrieval connects this spatial correspondence to planned actions. Think Twice retrieves camera and LiDAR features around coarse predictions and combines action-conditioned future scene prediction with cascaded refinement \cite{Jia2023ThinkTwice}. HiP-AD constructs planning queries at different spatial, temporal, and driving-style granularities and uses trajectory geometry to guide deformable attention within a unified planning--perception decoder \cite{tang2025hipad}. These methods establish the value of action-dependent feature selection. FeaXDrive2 further uses explicit vehicle-body geometry to distinguish body-corner query anchors from body-side visual sampling locations, with both determined by evolving trajectory estimates.

Vehicle-body occupancy also has a direct role in geometric trajectory optimization. Embodied Footprints encloses the swept occupancy between collocation points using enlarged footprints, allowing discrete collision constraints to account for vehicle motion between sampled poses \cite{li2024embodiedfootprints}. Choi et al. combine disc-based swept-volume estimation with B-spline trajectory optimization and incremental path flattening to reduce conservative occupancy approximations and support passage through narrow corridors \cite{choi2025sweptvolume}. For multi-axle vehicles, Li et al. integrate collision fields, planning corridors, and vehicle dynamics in multi-obstacle path optimization \cite{li2024multiaxle}. These studies explicitly relate body geometry and swept occupancy to collision avoidance and available passage space.

Learning-based planners incorporate environmental constraints through scene costs and generative guidance. ST-P3 combines predicted occupancy and map representations with planning costs \cite{hu2022stp3}, and VAD imposes constraints using vectorized roads and traffic participants \cite{jiang2023vad}. In generative planning, FeaXDrive applies adaptive curvature constraints and drivable-region signed-distance guidance at vehicle-body corners \cite{wang2026feaxdrive}; GuideFlow combines constraint guidance with energy-based models \cite{liu2026guideflow}; and Diffusion in Corridor incorporates spatiotemporal corridors and a differentiable vehicle simulator into denoising \cite{zhong2026diffusioncorridor}.

Together, these studies motivate connecting body-aware geometric planning with trajectory-conditioned visual retrieval. FeaXDrive2 uses vehicle-body geometry and evolving Flow trajectory estimates to select local environmental features, allowing the resulting body--scene context to condition subsequent trajectory generation.

\subsection{Safety Fine-Tuning of Pretrained Generative Planners}
\label{sec:II-C}

Safety fine-tuning of generative planners requires consideration of both training states and actual generated outcomes. For iterative generative models, DRaFT propagated differentiable rewards through the diffusion sampling process to directly optimize final outputs and investigated truncated backpropagation to reduce computational costs \cite{clark2024draft}. In driving, CARL-Diffusion uses simulator feedback to fine-tune candidate scoring through reinforcement learning while freezing its diffusion generator \cite{kang2026carldiffusion}. These studies apply feedback-driven optimization at different points in the planning pipeline. For Flow-based planners with trajectory-dependent spatial retrieval, supervision should further account for how intermediate trajectory estimates influence subsequent visual conditioning.

Targeted safety improvements in high-risk scenarios must also consider the preservation of established behaviors. Learning without Forgetting mitigated catastrophic forgetting through reference-output constraints \cite{li2018learningwithoutforgetting}. In autonomous driving, R2SE combined difficult-scenario identification, reinforcement learning for residual experts, and self-aware policy expansion to improve performance in challenging scenarios while maintaining general driving capabilities \cite{liu2026r2se}. Unlike reward-driven residual policy expansion, RC-F combines high-risk trajectory supervision using safety-validated demonstrations with reference behavior preservation using a frozen policy and matched initial noise. Its focus lies in jointly optimizing Flow generation with dynamic spatial retrieval through these complementary supervision objectives.

Overall, these studies motivate fine-tuning strategies that account for generated outcomes while preserving established behaviors. FeaXDrive2 adopts RC-F, combining high-risk trajectory supervision with reference behavior preservation under matched initial noise to optimize Flow-based planning with dynamic spatial retrieval.

Taken together, prior studies provide foundations for multimodal trajectory generation, spatially grounded visual interaction, and safety optimization. Building on these advances, FeaXDrive2 establishes dynamic body--scene spatial grounding, where vehicle-body geometry defines query anchors and evolving Flow trajectory estimates update visual retrieval locations. This mechanism, together with targeted safety optimization, enhances VESF.

\section{Problem Formulation and FeaXDrive2 Framework}
\label{sec:III}

\begin{figure*}[!t]
\centering
\includegraphics[width=0.95\textwidth]{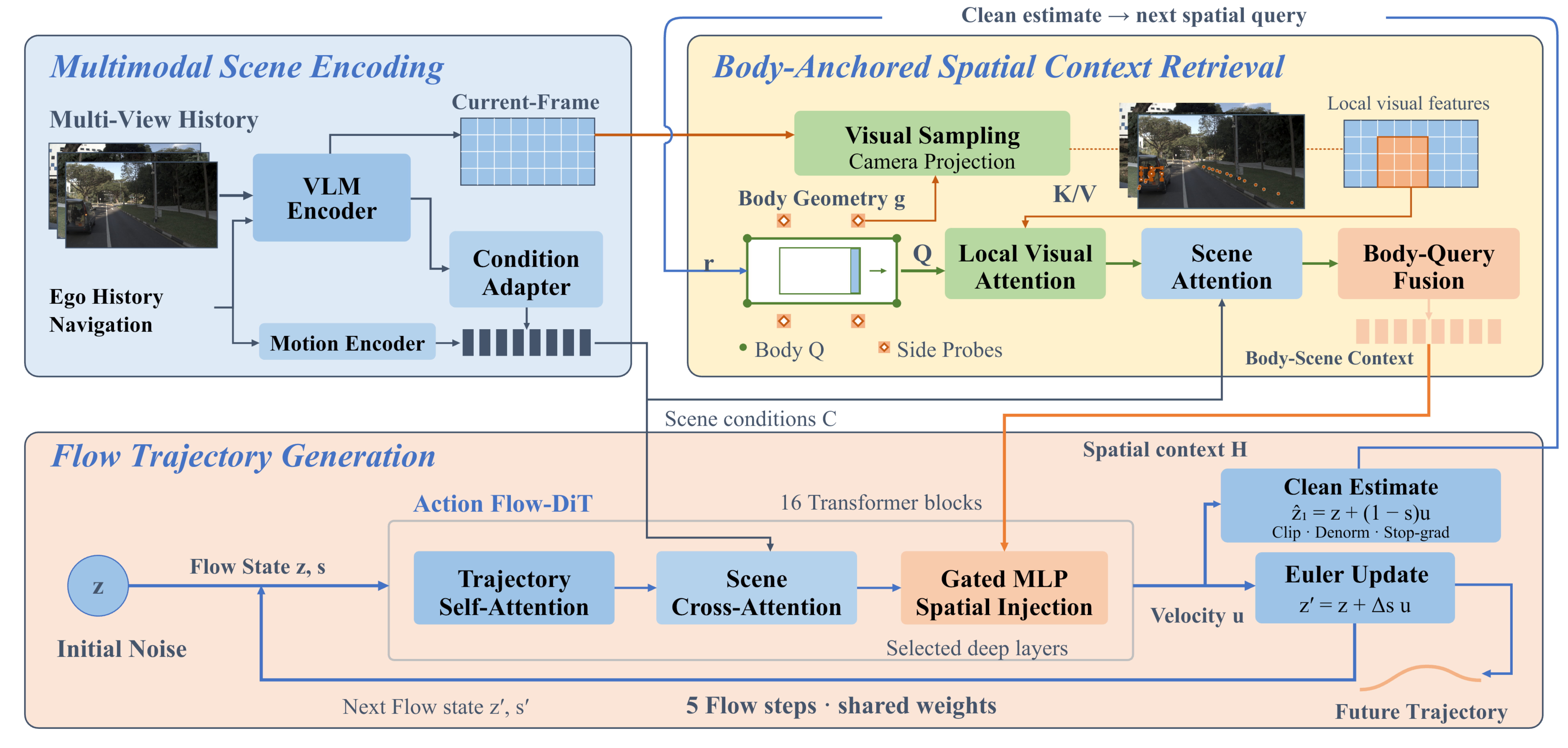}
\caption{Overall architecture of FeaXDrive2. Vehicle-body geometry and evolving trajectory estimates define body-anchored queries and side-probe locations for visual retrieval. The aggregated body--scene context conditions Action Flow-DiT velocity predictions. Euler integration updates the Flow state, while clean-trajectory estimates determine spatial queries for the next step.}
\label{fig:2}
\end{figure*}

As illustrated in Fig.~\ref{fig:2}, FeaXDrive2 is an end-to-end generative driving framework designed to enhance VESF. Conditioned on multimodal scene representations, the framework generates future trajectories using an Action Flow-DiT while incorporating body-anchored spatial visual information into the generation process. Training begins with base planner learning, followed by safety evaluation and vehicle-geometry counterfactual analysis to construct stratified training samples. The spatially enhanced planner is optimized through RC-F, followed by VLM--Action Alignment. This section formulates the planning task and VESF and presents the overall framework.

\subsection{Task Formulation and Vehicle-Embodied Spatial Feasibility}
\label{sec:III-A}

End-to-end driving planning aims to generate a future trajectory conditioned on multimodal observations, ego-motion states, and navigation information. Let \(O=(\{I_{c,\nu}\},s_{\mathrm{hist}},n)\) denote the input, where \(I_{c,\nu}\) is the image captured by camera \(c\) at historical time index \(\nu\), including the current frame; \(s_{\mathrm{hist}}\) represents the current and historical ego-motion states; and \(n\) denotes the navigation condition. The output is a trajectory \(\tau=(q_1,\ldots,q_H)\in\mathbb R^{H\times3}\) consisting of \(H\) future poses, where \(q_h=(x_h,y_h,\psi_h)\) specifies the position and heading at \(t_h=h\Delta t\), with \(\Delta t\) denoting the trajectory sampling interval. All poses are expressed in the current ego rear-axle coordinate frame. We use \(h\) to index future physical time and \(k\) to index Flow generation steps.

Given a vehicle geometry, VESF characterizes the extent to which the vehicle-body occupancy along a planned trajectory satisfies environmental constraints. Let \(g=[W,L_f,L_r]\) denote the vehicle geometry, where \(W\) is the vehicle width, and \(L_f\) and \(L_r\) are the distances from the rear axle to the front and rear ends, respectively. In the local rear-axle coordinate frame, the planar body footprint is represented by \(\mathcal B(g)=[-L_r,L_f]\times[-W/2,W/2]\). For a future pose \(q_h\), the corresponding body occupancy is

\begin{equation}
\label{eq:1}
\mathcal F(q_h,g)
=\left\{\begin{bmatrix}x_h\\y_h\end{bmatrix}
+R(\psi_h)b\;\middle|\;b\in\mathcal B(g)\right\},
\end{equation}

where \(R(\psi_h)\) is the planar rotation matrix and \(b\) denotes a point within the local body footprint. Given a traffic environment \(\mathcal E\), let \(\mathcal A_h\) and \(\mathcal D_h\) denote the obstacle occupancy and drivable region at future time \(t_h\), respectively. A spatially feasible trajectory should satisfy the following basic geometric constraints:

\begin{equation}
\label{eq:2}
\mathcal F(q_h,g)\cap\mathcal A_h=\varnothing,
\qquad
\mathcal F(q_h,g)\subseteq\mathcal D_h,
\qquad h=1,\ldots,H.
\end{equation}

These conditions characterize the spatial constraints between the vehicle and its surroundings. Even when the rear-axle reference point avoids obstacles, the full vehicle body may still violate spatial constraints. VESF is therefore jointly determined by vehicle geometry, trajectory poses, and environmental constraints.

\subsection{Base Planner Learning}
\label{sec:III-B}

Base planner learning establishes multimodal-conditioned continuous trajectory generation. The base planner consists of a pretrained vision-language encoder, a multimodal condition adapter, and an Action Flow-DiT. Alpamayo 1.5 \cite{nvidia2026alpamayo15} encodes multiview temporal images, ego-motion history, and navigation information, providing visual, state, navigation, and driving-reasoning representations. The condition adapter uses learnable queries to resample hidden representations from these sources into a fixed-length scene-conditioning sequence \(C_\eta(O)\), where \(\eta\) denotes the adapter parameters.

Action Flow-DiT comprises 16 Transformer blocks \cite{vaswani2017attention}. Trajectory self-attention layers alternate with scene-conditioned cross-attention layers, with the former modeling dependencies among future time steps and the latter integrating multimodal context. Specifically, trajectory features serve as queries, and \(C_\eta(O)\) provides the keys and values. Together with global scene conditioning, ego-state embeddings, and historical trajectory embeddings, the resulting representations form the conditioning input \(c_\eta(O)\) for velocity-field prediction.

The base planner employs Flow Matching to learn a continuous generative mapping from Gaussian noise to demonstrated trajectories. Given a demonstrated trajectory \(\tau^{\mathrm{GT}}\), its position and heading components are normalized using a fixed offset \(\mu\) and a positive diagonal scaling matrix \(S\), yielding \(z_1=\operatorname{Norm}(\tau^{\mathrm{GT}})\). The corresponding inverse transformation is denoted by \(\operatorname{Denorm}\). The same normalization parameters are used for base planner learning and subsequent RC-F.

We sample Gaussian noise \(z_0\sim\mathcal N(0,I)\) with the same dimensionality as \(z_1\), together with a generation time \(s\sim\mathcal U[0,1)\), and construct the interpolated state and target velocity:

\begin{equation}
\label{eq:3}
z_s=(1-s)z_0+sz_1,
\qquad u_s=z_1-z_0.
\end{equation}

The Action Flow-DiT, parameterized by \(\theta\), predicts a conditional velocity field \(v_\theta\) by minimizing

\begin{equation}
\label{eq:4}
\mathcal L_{\mathrm{FM}}
=\mathbb E_{(O,\tau^{\mathrm{GT}}),z_0,s}
\Big[\frac{1}{3H}
\big\|v_\theta(z_s,s,c_\eta(O))-(z_1-z_0)\big\|_F^2\Big].
\end{equation}

where \(\|\cdot\|_F\) denotes the Frobenius norm. During base planner learning, the vision-language encoder is frozen, while the multimodal condition adapter and Action Flow-DiT are optimized using \(\mathcal L_{\mathrm{FM}}\).

At inference time, the base planner starts from \(z^{(0)}=z_0\) and performs \(K=5\) Euler updates at generation times \(s_k=k/K\). At each step, the planner evaluates the predicted velocity field at the current state \(z^{(k)}\), scales the resulting velocity by the step size \(\Delta s=1/K\), and adds it to the current state. The final state \(z^{(K)}\) is clipped to \([-1,1]\) and denormalized to obtain the future trajectory. Predictions from the resulting base policy \(\pi_{\mathrm{base}}\) are used for subsequent safety evaluation and vehicle-geometry counterfactual analysis. The base policy also provides the initial parameters and reference behavior for RC-F.

\subsection{VESF-Aware Spatial Modeling and Safety Adaptation}
\label{sec:III-C}

Body-Anchored Spatial Context Retrieval conditions the base planner on local visual information selected using vehicle geometry and evolving trajectory estimates. RC-F optimizes the spatially enhanced planner through high-risk trajectory supervision and reference behavior preservation. VLM--Action Alignment then jointly adapts multimodal conditioning and selected Action Flow-DiT parameters through planning supervision. The complete planner is formulated as \(\tau_{\theta,\phi,\omega}=\pi_{\theta,\phi,\omega}(O,g,\Gamma;z_0)\). Here, \(\theta\), \(\phi\), and \(\omega\) denote the Action Flow-DiT, spatial retrieval and gated injection, and language-model LoRA parameters, respectively; \(\Gamma\) contains camera calibration and image-processing parameters. Section~\ref{sec:IV} details the spatial retrieval and optimization mechanisms.

\section{VESF-Aware Spatial Modeling and Safety Optimization}
\label{sec:IV}

This section presents body--scene spatial grounding and RC-F to enhance VESF, followed by VLM--Action Alignment to further optimize the resulting planner. Vehicle geometry and evolving trajectory estimates jointly determine spatial queries, while the retrieved body--scene spatial context conditions subsequent Flow updates. RC-F further optimizes generated trajectories through high-risk supervision and reference behavior preservation.

\subsection{Body-Anchored Spatial Query Construction}
\label{sec:IV-A}

Vehicle geometry defines body-corner query anchors, while evolving trajectory estimates determine their spatial positions during Flow generation. Let \(r^{(k)}=(r_1^{(k)},\ldots,r_H^{(k)})\) denote the trajectory estimate in metric coordinates available at the beginning of Flow step \(k\), where \(r_h^{(k)}=(x_h^{(k)},y_h^{(k)},\psi_h^{(k)})\). Given vehicle geometry \(g=[W,L_f,L_r]\), the front and rear corners are located at \((L_f,\pm W/2)\) and \((-L_r,\pm W/2)\), respectively, in the local rear-axle coordinate frame. Let \(o_j^{\mathrm{body}}\), \(j\in\{1,2,3,4\}\), denote these local offsets, ordered as front-left, front-right, rear-left, and rear-right. Transforming the offsets according to each estimated trajectory pose yields the corresponding anchors in the current ego coordinate frame:

\begin{equation}
\label{eq:5}
p_{h,j}^{\mathrm{body},(k)}
=\begin{bmatrix}x_h^{(k)}\\y_h^{(k)}\end{bmatrix}
+R(\psi_h^{(k)})o_j^{\mathrm{body}},
\qquad j=1,\ldots,4.
\end{equation}

To encode vehicle dimensions, spatial position, and generation state, a nine-dimensional input is constructed for each corner:

\begin{equation}
\label{eq:6}
\begin{aligned}
a_{h,j}^{(k)}
&=\big[
W,L_f,L_r,\,
p_{h,j,x}^{\mathrm{body},(k)},p_{h,j,y}^{\mathrm{body},(k)},\\
&\qquad
\sin\psi_h^{(k)},\cos\psi_h^{(k)},\,
s_k,h/H
\big],\\
Q_{h,j}^{(k)}
&=\operatorname{MLP}_{q}(a_{h,j}^{(k)})+e_j^{\mathrm{corner}}.
\end{aligned}
\end{equation}

Here, \(p_{h,j,x}^{\mathrm{body},(k)}\) and \(p_{h,j,y}^{\mathrm{body},(k)}\) denote the coordinates of the body-corner anchor, while \(s_k\) and \(h/H\) indicate the Flow generation time and normalized prediction time, respectively. The query encoder \(\operatorname{MLP}_q\) maps the nine-dimensional input to feature dimension \(D\), and \(e_j^{\mathrm{corner}}\) is a learnable corner-type embedding. Each future pose produces four queries, yielding \(4H\) body-anchored queries for the entire trajectory.

\subsection{Trajectory-Conditioned Spatial Visual Retrieval}
\label{sec:IV-B}

Trajectory-conditioned spatial visual retrieval uses body-anchored queries to aggregate local visual features sampled at body-side probe locations, forming body--scene spatial context. The resulting context is injected into Action Flow-DiT through gated residual connections, while the retrieval locations are updated using preceding clean-trajectory estimates across Flow steps.

Let \(L=L_f+L_r\) denote the vehicle length, and \(d\) the lateral offset of the probe locations from the corresponding body sides. Following the same ordering as the four body-corner queries, let \(o_j^{\mathrm{side}}\in\mathbb R^2\) denote the local coordinates of side probe \(j\) relative to the rear axle:

\begin{equation}
\label{eq:7}
\begin{aligned}
o_1^{\mathrm{side}}&=(-L_r+0.75L,\ W/2+d),\\
o_2^{\mathrm{side}}&=(-L_r+0.75L,\ -W/2-d),\\
o_3^{\mathrm{side}}&=(-L_r+0.25L,\ W/2+d),\\
o_4^{\mathrm{side}}&=(-L_r+0.25L,\ -W/2-d).
\end{aligned}
\end{equation}

Applying the same pose transformation as for the body corners yields the visual retrieval locations \(p_{h,j}^{\mathrm{side},(k)}\) in the current ego coordinate frame. The offset \(d\) specifies the sampling layout rather than a measured obstacle clearance or safety threshold.

Visual retrieval operates on current multiview spatial features \(F_c(O)\) extracted from the pretrained vision-language encoder before language fusion. Camera features are resampled to a common spatial grid. Using a local horizontal plane passing through the current rear-axle origin as the projection reference, each probe is represented as \(\bar p_{h,j}^{\mathrm{side},(k)}=(p_{h,j,x}^{\mathrm{side},(k)},p_{h,j,y}^{\mathrm{side},(k)},0)\). The probe is projected into image-feature coordinates using camera calibration and image-processing parameters \(\Gamma_c\):

\begin{equation}
\label{eq:8}
\xi_{c,h,j}^{(k)}
=\mathcal S_c\!\left(\Pi_{\Gamma_c}
\left(\bar p_{h,j}^{\mathrm{side},(k)}\right)\right),
\end{equation}

where \(\Pi_{\Gamma_c}\) accounts for camera projection, lens distortion, and image preprocessing transformations, while \(\mathcal S_c\) converts pixel coordinates into continuous feature coordinates \(\xi_{c,h,j}^{(k)}\). The horizontal plane serves as a reference for visual retrieval rather than representing measured road surfaces or obstacle geometry. Bilinear sampling is performed within a local neighborhood centered at each valid projected location. The sampled features are linearly projected to the query dimension and augmented with camera-source and neighborhood-offset embeddings. Invalid projections are masked during attention, and a learnable null-visual token is used when all projections are invalid.

The local features associated with each query form a key--value sequence \(V_{h,j}^{(k)}\), with validity mask \(M_{h,j}^{(k)}\). Each body-anchored query first aggregates local visual features and then interacts with the multimodal scene-conditioning sequence \(C_\eta(O)\). Features from the four corners at each future time step are subsequently fused:

\begin{equation}
\label{eq:9}
\begin{aligned}
\widetilde Q_{h,j}^{(k)}
&=Q_{h,j}^{(k)}
+\operatorname{MHA}_{v}\!
\left(\operatorname{LN}_{v}(Q_{h,j}^{(k)}),
V_{h,j}^{(k)},V_{h,j}^{(k)};\,M_{h,j}^{(k)}\right),\\
\overline Q_{h,j}^{(k)}
&=\widetilde Q_{h,j}^{(k)}
+\operatorname{MHA}_{s}\!
\left(\operatorname{LN}_{s}(\widetilde Q_{h,j}^{(k)}),
C_\eta(O),C_\eta(O)\right),\\
H_h^{(k)}
&=W_p\big[
\overline Q_{h,1}^{(k)};\overline Q_{h,2}^{(k)};
\overline Q_{h,3}^{(k)};\overline Q_{h,4}^{(k)}
\big]+b_p.
\end{aligned}
\end{equation}

Here, \(\operatorname{MHA}_v\) and \(\operatorname{MHA}_s\) denote the multi-head attention modules for local visual features and scene conditioning, respectively. The updated queries after these two operations are denoted by \(\widetilde Q\) and \(\overline Q\), respectively, and \(\operatorname{LN}\) denotes layer normalization. The notation \([\,;\,]\) denotes feature concatenation. The weight matrix \(W_p\) and bias \(b_p\) are used to fuse the concatenated features from the four corners into \(H_h^{(k)}\). Stacking \(H_h^{(k)}\) across the \(H\) future time steps yields the Body--Scene Spatial Context Representation \(H_{\mathrm{body\text{-}scene}}^{(k)}\in\mathbb R^{H\times D}\).

The resulting spatial context is incorporated into Action Flow-DiT through gated residual connections. Let \(A_l^{(k)}\in\mathbb R^{H\times D}\) denote trajectory features at injection layer \(l\) during generation step \(k\). The update is given by

\begin{equation}
\label{eq:10}
\begin{aligned}
A_l^{(k)}\leftarrow{}& A_l^{(k)}+\tanh(\alpha_l)\,
\operatorname{MLP}_{l}\Big(
\Big[\operatorname{LN}_{a,l}(A_l^{(k)});\\
&\qquad\operatorname{LN}_{b,l}(H_{\mathrm{body\text{-}scene}}^{(k)})\Big]
\Big).
\end{aligned}
\end{equation}

The two layer normalization operations separately normalize the trajectory features and body--scene spatial context, while \(\operatorname{MLP}_l\) maps their concatenated features at each future time step. The learnable scalar \(\alpha_l\) controls the residual magnitude through \(\tanh(\alpha_l)\). Initializing \(\alpha_l\) to zero ensures that the added spatial branch does not alter the base planner's output at the start of training. The gated residual subsequently learns to incorporate the retrieved spatial context into trajectory generation.

Spatial retrieval and trajectory generation are coupled through clean-trajectory estimates updated across Flow steps. As the estimated trajectory changes in position and heading, the body-corner anchors and side-probe locations are updated accordingly, allowing visual information retrieval to adapt to the evolving trajectory. Let \(F(O)=\{F_c(O)\}\) and \(\Gamma=\{\Gamma_c\}\). The parameter set \(\phi\) includes spatial retrieval and gated-injection parameters, while the retrieval operator \(\mathcal R_\phi\) depends only on the retrieval subset. At generation step \(k\), spatial context is constructed from the available trajectory estimate \(r^{(k)}\), followed by velocity-field prediction and trajectory updates:

\begin{equation}
\label{eq:11}
\begin{aligned}
H_{\mathrm{body\text{-}scene}}^{(k)}
&=\mathcal R_\phi
\big(F(O),C_\eta(O),g,r^{(k)},s_k;\Gamma\big),\\
u^{(k)}
&=v_{\theta,\phi}
\big(z^{(k)},s_k,c_\eta(O),
H_{\mathrm{body\text{-}scene}}^{(k)}\big),\\
\hat z_1^{(k)}
&=z^{(k)}+(1-s_k)u^{(k)},\\
z^{(k+1)}
&=z^{(k)}+\Delta s\,u^{(k)},\\
r^{(k+1)}
&=\operatorname{sg}\left[
\operatorname{Denorm}\left(
\operatorname{Clip}_{[-1,1]}(\hat z_1^{(k)})
\right)\right].
\end{aligned}
\end{equation}

Here, \(u^{(k)}\) is the predicted velocity field, \(\hat z_1^{(k)}\) is the clean-trajectory estimate extrapolated from the current Flow state, and \(\operatorname{Clip}_{[-1,1]}\) denotes element-wise clipping. Generation starts from \(z^{(0)}=z_0\), with the metric query trajectory initialized as an all-zero position and heading sequence, \(r^{(0)}=\mathbf0\). At subsequent steps, spatial queries are recomputed from the preceding clean-trajectory estimate. The final planned trajectory is

\begin{equation}
\label{eq:12}
\tau_{\theta,\phi}
=
\operatorname{Denorm}\left(
\operatorname{Clip}_{[-1,1]}(z^{(K)})
\right).
\end{equation}

The stop-gradient operator \(\operatorname{sg}\) prevents gradients from propagating to preceding Flow states through trajectory-dependent spatial query updates, avoiding additional cross-step gradient paths involving geometric projection and visual feature sampling. During RC-F, the spatial retrieval parameters remain trainable through context injection, and the Euler state recurrence \(z^{(k)}\rightarrow z^{(k+1)}\) remains differentiable. Since training and inference follow the same five-step Euler process, final-trajectory losses can still optimize the planner through the actual generation chain.

\subsection{Rollout-Consistent Safety Fine-Tuning}
\label{sec:IV-C}

\begin{figure}[!t]
\centering
\includegraphics[width=\columnwidth]{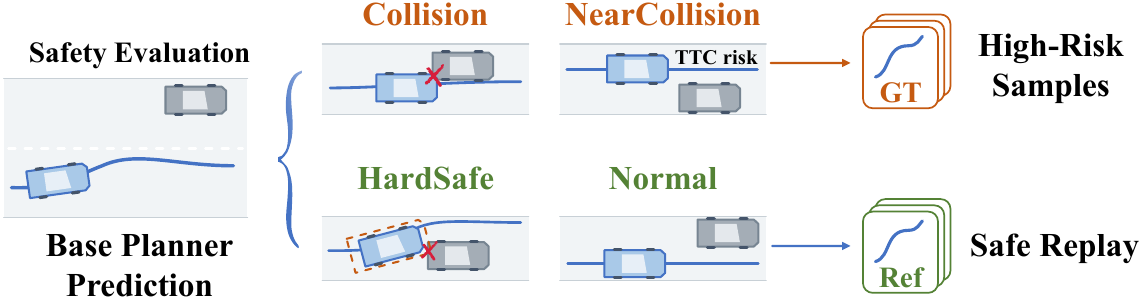}
\caption{Training sample construction for RC-F. Base-planner safety evaluation identifies Collision and NearCollision samples for high-risk supervision, while body-clearance and vehicle-geometry counterfactual analyses distinguish HardSafe from Normal for reference behavior preservation. The dashed outline indicates the larger counterfactual footprint.}
\label{fig:3}
\end{figure}

RC-F combines high-risk trajectory supervision with reference behavior preservation to optimize multi-step Flow generation. As illustrated in Fig.~\ref{fig:3}, safety evaluation of the base planner identifies collision and near-collision samples, which use safety- and progress-validated demonstrations. Among base-safe samples without geometric contact, vehicle-geometry counterfactual analysis and minimum body clearance distinguish HardSafe from Normal. Both safe subsets share the same reference behavior preservation objective and use the frozen base planner as the reference policy \(\pi_{\mathrm{ref}}=\pi_{\mathrm{base}}\). HardSafe samples receive a higher sampling weight relative to their prevalence in the training pool to preserve safe behavior in spatially constrained or geometry-sensitive scenes, while Normal samples maintain existing safe behavior across broader driving scenarios.

For each training sample, the trainable planner executes the complete five-step Flow rollout from initial noise \(z_0\), producing intermediate states \(z^{(k)}\), clean-trajectory estimates \(\hat z_1^{(k)}\), spatial query conditions \(r^{(k)}\), and the final trajectory \(\tau_{\theta,\phi}\). For behavior preservation samples, the frozen reference policy independently generates states \(z_{\mathrm{ref}}^{(k)}\) and trajectory \(\tau_{\mathrm{ref}}\) under the same observation and initial noise. These rollouts provide the intermediate and final outputs for safety optimization.

Auxiliary local objectives complement rollout supervision: \(\ell_{\mathrm{FM}}\) provides local Flow Matching supervision for high-risk samples, while \(\ell_{\mathrm{velocity}}\) matches frozen reference velocities for behavior preservation samples. Both are evaluated at constructed interpolation states; the trainable planner constructs their spatial context from trajectory estimates generated at preceding Flow steps. In contrast, trajectory and state objectives act on the actual rollout. Appendix~\ref{app:A} specifies the interpolation construction and auxiliary losses.

For trajectory supervision, \(\rho\) denotes a coordinate-scaled Smooth L1 loss, and \(e(\tau,\tau')\) the position and heading error between two trajectories. The scaling matrix \(S\), defined in Section~\ref{sec:III-B}, converts normalized state differences to physical units. Error scales and angular-error handling are specified in Appendix~\ref{app:A}.

High-risk trajectory supervision combines local velocity supervision with losses on intermediate clean-trajectory estimates and the final generated trajectory. Let \(z^{\mathrm{GT}}=\operatorname{Norm}(\tau^{\mathrm{GT}})\), and let \(w_k\) denote nonnegative weights that increase with Flow generation progress and sum to one. Since \(\hat z_1^{(K-1)}=z^{(K)}\) under uniform-step Euler integration, the intermediate loss is applied only to the first \(K-1\) clean-trajectory estimates, while the final output is supervised separately:

\begin{equation}
\label{eq:13}
\begin{aligned}
\ell_{\mathrm{clean}}
&=\sum_{k=0}^{K-2}w_k\,
\rho\left((\hat z_1^{(k)}-z^{\mathrm{GT}})S\right),\\
\ell_{\mathrm{final,GT}}
&=\rho\left(e(\tau_{\theta,\phi},\tau^{\mathrm{GT}})\right),\\
\ell_{\mathrm{HR}}
&=0.1\,\ell_{\mathrm{FM}}
+0.5\,\ell_{\mathrm{clean}}
+\ell_{\mathrm{final,GT}}.
\end{aligned}
\end{equation}

Reference behavior preservation constrains the independently evolving trainable and reference policies under shared observations and initial noise. Intermediate states are compared after the first \(K-1\) Euler updates, while the final trajectories receive a separate preservation loss:

\begin{equation}
\label{eq:14}
\begin{aligned}
\ell_{\mathrm{state}}
&=\frac{1}{K-1}\sum_{k=1}^{K-1}
\rho\left((z^{(k)}-z_{\mathrm{ref}}^{(k)})S\right),\\
\ell_{\mathrm{final,ref}}
&=\rho\left(e(\tau_{\theta,\phi},\tau_{\mathrm{ref}})\right),\\
\ell_{\mathrm{BP}}
&=0.1\,\ell_{\mathrm{velocity}}
+0.25\,\ell_{\mathrm{state}}
+\ell_{\mathrm{final,ref}}.
\end{aligned}
\end{equation}

The two objectives are combined according to sample role. Let \(\mathbb I_{\mathrm{HR}}\) and \(\mathbb I_{\mathrm{BP}}\) indicate high-risk supervision and behavior preservation samples, respectively. The joint objective is

\begin{equation}
\label{eq:15}
\mathcal L_{\mathrm{RCF}}
=\mathbb E\left[
\mathbb I_{\mathrm{HR}}\ell_{\mathrm{HR}}
+\mathbb I_{\mathrm{BP}}\ell_{\mathrm{BP}}
\right].
\end{equation}

Here, the expectation is taken over stratified training samples, Gaussian initial noise, and uniformly sampled discrete Flow steps for auxiliary supervision. During RC-F, the trainable planner is initialized from the base planner, and only the spatial retrieval modules and selected deeper Action Flow-DiT parameters are optimized, while the remaining pretrained parameters and the reference policy are frozen.

\textbf{VLM--Action Alignment.} Starting from the Body--Scene Grounded Planner, we jointly update language-model LoRA parameters and selected Action Flow-DiT parameters through planning supervision. The visual encoder, condition adapter, spatial retrieval module, and remaining planner parameters are frozen. The objective retains high-risk trajectory supervision and downweights reference behavior preservation, using the pre-alignment planner as the frozen reference policy. Planning gradients propagate through the five-step rollout to both parameter groups.

In summary, FeaXDrive2 combines dynamic body--scene spatial grounding and targeted safety optimization within Flow-based trajectory generation. Spatial retrieval acquires body-relevant environmental information using evolving trajectory estimates, while RC-F enhances the safety of generated trajectories through high-risk supervision and reference behavior preservation.

\section{Experiments}
\label{sec:V}

This section first evaluates overall planning performance on NAVSIM, followed by ablation studies examining the contributions of spatial retrieval and RC-F. Vehicle-geometry counterfactual analysis, body-clearance evaluation, and qualitative case studies are then used to investigate VESF-related planning behavior.

\subsection{Experimental Setup}
\label{sec:V-A}

\textbf{Datasets and evaluation protocols.} We conduct experiments on NAVSIM \cite{dauner2024navsim}, whose TRAIN, VAL, and TEST splits contain 85,109, 18,179, and 12,146 scenarios, respectively. NAVTRAIN is used for training and safety adaptation, the validation set is used for model selection, and NAVTEST is used for final evaluation. NAVSIM v1 evaluates planning performance using PDMS. For the extended evaluation, we employ the official single-stage scorer of NAVSIM v2 \cite{cao2025pseudosimulation} to compute EPDMS on the same NAVTEST scenarios.

\textbf{Evaluation metrics.} For NAVSIM v1, we report the overall planning score (PDMS), no-at-fault collision (NC), time-to-collision safety (TTC), drivable-area compliance (DAC), ego progress (EP), and comfort (C). NAVSIM v2 additionally reports driving-direction compliance (DDC), traffic-light compliance (TLC), lane keeping (LK), historical comfort (HC), and extended comfort (EC), together with EPDMS. All reported scores are obtained by averaging the per-scenario scores and multiplying by 100.

For safety analysis, we distinguish the mean NC score, at-fault collision events, and raw geometric contacts (Raw Contact). At-fault collision counts scenarios with an unscaled NC score below 1, whereas Raw Contact counts scenarios in which the vehicle body geometrically contacts an obstacle, regardless of collision responsibility. Because NC can take values of 0, 0.5, or 1, its mean is not equivalent to the fraction of scenarios without an at-fault collision. A joint safety failure is defined as \((NC<1)\lor(DAC<1)\lor(TTC<1)\); Raw Contact is not included as an additional criterion. We also report minimum body clearance, raw progress, and stopping rate for behavioral analysis.

\textbf{Training data construction.} RC-F constructs four categories of training samples comprising 82,678 scenarios, based on the base planner's safety evaluation and vehicle-geometry counterfactual analysis on NAVTRAIN (Table~\ref{tab:I}). Collision and NearCollision correspond to at-fault collision scenarios and contact-free TTC failure scenarios, respectively. For both categories, high-risk trajectory supervision uses demonstration trajectories that satisfy safety and progress criteria.

HardSafe samples are selected from scenarios in which the base planner is jointly safe and has no geometric contact. They include scenarios with minimum body clearance in \((0,0.5]\,\mathrm m\), as well as scenarios that are jointly safe under Pacifica geometry but exhibit a strict safety failure when the geometry is replaced with Large. The remaining eligible safe scenarios are assigned to Normal. HardSafe and Normal share the same reference behavior preservation objective, but HardSafe receives a higher sampling weight relative to its proportion in the dataset to strengthen preservation of safe behavior in spatially constrained scenarios. Table~\ref{tab:I} reports the sample counts and sampling proportions; complete eligibility criteria and exclusion rules are given in Appendix~\ref{app:B}.

\begin{table}[!tbp]
\caption{RC-F training data for high-risk trajectory supervision and reference behavior preservation.}
\label{tab:I}
\centering
\setlength{\tabcolsep}{3pt}
\renewcommand{\arraystretch}{1.15}
\begin{tabularx}{\linewidth}{@{}>{\raggedright\arraybackslash}p{.20\linewidth} >{\raggedleft\arraybackslash}p{.16\linewidth} >{\raggedright\arraybackslash}p{.31\linewidth} >{\raggedleft\arraybackslash}X@{}}
\toprule
Sample category & No. of scenarios & Supervision source & RC-F sampling proportion \\
\midrule
Collision & 369 & Safety-validated GT & 25\% \\
NearCollision & 1,753 & Safety-validated GT & 15\% \\
HardSafe & 8,673 & Frozen reference policy & 25\% \\
Normal & 71,883 & Frozen reference policy & 35\% \\
Total & 82,678 & --- & 100\% \\
\bottomrule
\end{tabularx}
\end{table}

\textbf{Model configuration and pretraining source.} FeaXDrive2 uses camera-only perception inputs, combining images from three forward-facing cameras with ego-motion states and navigation information to generate a single trajectory. The main experiments adopt Pacifica vehicle geometry and generate eight future trajectory poses over a 4s horizon through five Euler updates. The vision-language backbone is initialized from publicly available Alpamayo 1.5 pretrained weights \cite{nvidia2026alpamayo15}. All internal component comparisons use the same pretrained initialization.

\textbf{Training procedure and freezing strategy.} During base planner learning, the VLM is frozen, while the condition adapter and Action Flow-DiT are trained with Flow Matching for 100 epochs. RC-F is initialized from the base planner and optimizes the spatial retrieval module and selected deeper Action Flow-DiT parameters, while freezing the VLM, condition adapter, and remaining planner parameters. The frozen base policy provides reference behavior preservation supervision. Optimization uses AdamW \cite{loshchilov2019adamw} and a 4,000-step learning-rate schedule.

VLM--Action Alignment continues training from the Body--Scene Grounded Planner for 2,200 steps. It jointly optimizes language-model LoRA parameters and selected Action Flow-DiT parameters, while freezing the visual encoder, condition adapter, spatial retrieval module, and remaining parameters. The pre-alignment planner serves as the frozen reference policy, and planning supervision jointly adapts multimodal conditioning and action generation. Detailed layer indices, LoRA configurations, learning rates, and optimizer settings are provided in Appendix~\ref{app:C}.

\textbf{Model selection.} RC-F and VLM--Action Alignment use a fixed validation subset of 4,231 scenarios, consisting of 231 scenarios with at-fault collisions under the base planner, 1,000 HardSafe scenarios, and 3,000 Normal scenarios. For RC-F, model selection primarily considers the net reduction in at-fault collisions relative to the base planner, while also examining the preservation of safe behavior and driving progress. For VLM--Action Alignment, checkpoints are selected according to validation PDMS. The selected checkpoints are then fixed for NAVTEST evaluation.

\textbf{Comparison methods.} Results for publicly reported baselines are taken from their corresponding configurations in the literature. Internal ablations compare the Base Planner, RC-F w/o Spatial Retrieval, Body--Scene Grounded Planner, and the complete FeaXDrive2 under identical test scenarios, initial noise, and scoring settings. The spatial retrieval and RC-F mechanism analyses use the Body--Scene Grounded Planner as their reference, while the complete FeaXDrive2 is used to report final planning performance.

\subsection{Comparison with Existing Methods}
\label{sec:V-B}

\textbf{NAVSIM v1.} FeaXDrive2 achieves a PDMS of 88.8 with single-candidate inference. Its NC and TTC scores reach 99.1 and 96.3, respectively, the highest among the methods listed in Table~\ref{tab:II}. Compared with DriveDreamer-Policy, which achieves the highest PDMS in the table, FeaXDrive2 obtains NC and TTC scores that are 0.7 and 1.2 points higher, respectively.

\begin{table}[!tbp]
\caption{Planning performance on NAVSIM v1 NAVTEST.}
\label{tab:II}
\centering
\setlength{\tabcolsep}{3pt}
\renewcommand{\arraystretch}{1.15}
\begin{tabularx}{\linewidth}{@{}>{\raggedright\arraybackslash}Xrrrrr@{}}
\toprule
Method & NC \ensuremath{\uparrow} & TTC \ensuremath{\uparrow} & DAC \ensuremath{\uparrow} & EP \ensuremath{\uparrow} & PDMS \ensuremath{\uparrow} \\
\midrule
TransFuser \cite{Chitta2023TransFuser} & 97.7 & 92.8 & 92.8 & 79.2 & 84.0 \\
UniAD \cite{hu2023uniad} & 97.8 & 92.9 & 91.9 & 78.8 & 83.4 \\
TrajDiff \cite{gui2025trajdiff} & 98.1 & 94.2 & 95.9 & 81.7 & 87.5 \\
ReCogDrive-IL \cite{li2026recogdrive} & 98.1 & 94.2 & 94.7 & 80.9 & 86.5 \\
DiffusionDrive \cite{liao2025diffusiondrive} & 98.2 & 94.7 & 96.2 & 82.2 & 88.1 \\
BridgeDrive \cite{liu2026bridgedrive} & 98.2 & 94.5 & 96.1 & 82.3 & 88.0 \\
ARTEMIS \cite{feng2025artemis} & 98.3 & 94.3 & 95.1 & 81.4 & 87.0 \\
DriveDreamer-Policy \cite{zhou2026drivedreamerpolicy} & 98.4 & 95.1 & \textbf{97.1} & \textbf{83.5} & \textbf{89.2} \\
WoTE \cite{li2025wote} & 98.5 & 94.9 & \underline{96.8} & 81.9 & 88.3 \\
DriveVLA-W0 \cite{li2026drivevlaw0} & \underline{98.7} & \underline{95.5} & 96.2 & 82.2 & 88.4 \\
\textbf{FeaXDrive2} & \textbf{99.1} & \textbf{96.3} & 96.1 & \underline{82.4} & \underline{88.8} \\
\bottomrule
\end{tabularx}
\par\vspace{2pt}
\begin{minipage}{\linewidth}
\footnotesize\raggedright
\emph{Note:} Best and second-best results are shown in \textbf{bold} and \underline{underlined}, respectively. DriveVLA-W0 denotes the Query-based Action Expert variant.
\end{minipage}
\end{table}

\textbf{NAVSIM v2.} Under the single-stage non-reactive evaluation on the same NAVTEST scenarios, FeaXDrive2 achieves an EPDMS of 88.3. Its NC and TTC scores are 99.1 and 98.6, respectively, both the highest among the methods listed in Table~\ref{tab:III}.

\begin{table*}[!tbp]
\caption{Reported EPDMS on NAVTEST and evaluation with the NAVSIM v2 scorer.}
\label{tab:III}
\centering
\setlength{\tabcolsep}{3pt}
\renewcommand{\arraystretch}{1.15}
\begin{tabularx}{\linewidth}{@{}>{\raggedright\arraybackslash}Xrrrrrrrrrr@{}}
\toprule
Method & NC \ensuremath{\uparrow} & TTC \ensuremath{\uparrow} & DAC \ensuremath{\uparrow} & DDC \ensuremath{\uparrow} & TLC \ensuremath{\uparrow} & EP \ensuremath{\uparrow} & LK \ensuremath{\uparrow} & HC \ensuremath{\uparrow} & EC \ensuremath{\uparrow} & EPDMS \ensuremath{\uparrow} \\
\midrule
TransFuser \cite{Chitta2023TransFuser} & 96.9 & 95.4 & 89.9 & 97.8 & 99.7 & 87.1 & 92.7 & \textbf{98.3} & \textbf{87.2} & 76.7 \\
DriveDreamer-Policy \cite{zhou2026drivedreamerpolicy} & 98.4 & 97.7 & 97.1 & \underline{99.5} & \textbf{99.9} & \underline{87.9} & \textbf{97.6} & \textbf{98.3} & 79.4 & \textbf{88.7} \\
DriveSuprim \cite{yao2025drivesuprim} & 98.4 & 97.8 & \underline{98.6} & \textbf{99.6} & \underline{99.8} & \textbf{90.5} & 97.0 & \textbf{98.3} & 78.6 & 87.1 \\
DriveVLA-W0 \cite{li2026drivevlaw0} & \underline{98.5} & \underline{98.1} & \textbf{99.1} & 98.0 & 99.7 & 86.4 & 93.2 & 97.9 & 58.9 & 86.1 \\
\textbf{FeaXDrive2} & \textbf{99.1} & \textbf{98.6} & 96.1 & \underline{99.5} & \textbf{99.9} & 86.9 & \underline{97.3} & \underline{98.2} & \underline{80.2} & \underline{88.3} \\
\bottomrule
\end{tabularx}
\end{table*}

FeaXDrive2 achieves competitive overall planning performance while attaining strong collision-safety and TTC results among the compared methods.

\subsection{Ablation Studies}
\label{sec:V-C}

\subsubsection{Effect of Core Components}

\begin{figure}[!t]
\centering
\includegraphics[width=\columnwidth]{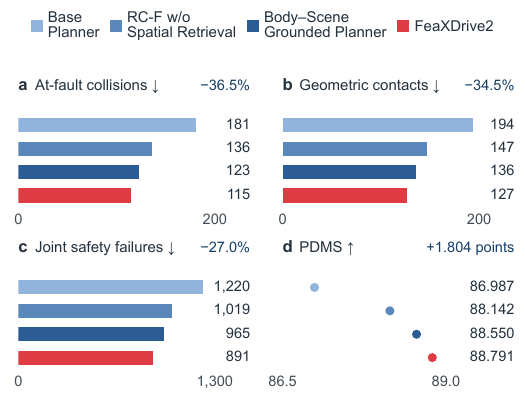}
\caption{Planning performance across planner configurations on NAVTEST. All 12,146 scenarios are included. (a) At-fault collisions; (b) geometric contacts; (c) joint safety failures; and (d) PDMS. Annotations indicate changes relative to the Base Planner, expressed as percentages in (a--c) and PDMS points in (d).}
\label{fig:4}
\end{figure}

The progressive comparison of component configurations demonstrates incremental planning improvements from RC-F, body--scene spatial grounding, and VLM--Action Alignment (Fig.~\ref{fig:4}). RC-F w/o Spatial Retrieval increases the Base Planner's PDMS from 86.987 to 88.142 and reduces the number of at-fault collision scenarios from 181 to 136, a 24.9\% reduction. These results indicate that combining high-risk trajectory supervision with reference behavior preservation improves the safety and overall planning performance of the base planner.

Body--scene spatial grounding further improves planning safety on top of RC-F. Compared with RC-F w/o Spatial Retrieval, the Body--Scene Grounded Planner improves PDMS by 0.408 points, reduces at-fault collision scenarios by 13 (9.6\%), and reduces joint safety failures by 54 (5.3\%). This comparison supports the incremental value of introducing body-relevant spatial context into the safety-adapted planner.

VLM--Action Alignment provides further gains for the body--scene grounded planner. Relative to the Body--Scene Grounded Planner, the complete FeaXDrive2 improves PDMS by 0.241 points, reduces at-fault collision scenarios by 8 (6.5\%), and reduces joint safety failures by 74 (7.7\%). This result indicates that jointly adapting multimodal scene conditioning and action generation through planning supervision can further improve planning performance beyond spatial grounding and safety optimization.

\subsubsection{Spatial Grounding}

The spatial retrieval ablations use the Body--Scene Grounded Planner as the reference to examine body-anchored queries, side-probe sampling, cross-step query updates, and local visual output. All retrieval configurations retain four query slots and the same context-injection structure, with their query references and sampling layouts summarized in Table~\ref{tab:IV}(a).

\begin{table*}[!tbp]
\caption{Ablation of spatial grounding on NAVTEST.}
\label{tab:IV}
\centering
\setlength{\tabcolsep}{3pt}
\renewcommand{\arraystretch}{1.15}
\textit{(a) Spatial retrieval configurations.}\par\smallskip
\begin{tabularx}{\linewidth}{@{}l l l >{\centering\arraybackslash}X >{\centering\arraybackslash}X r@{}}
\toprule
Method & Query geometric reference & Visual sampling location & At-fault collisions~\ensuremath{\downarrow} & Geometric contacts~\ensuremath{\downarrow} & PDMS \ensuremath{\uparrow} \\
\midrule
RC-F w/o Spatial Retrieval & --- & --- & 136 & 147 & 88.142 \\
Centerline Retrieval & Trajectory reference point & Centerline & 137 & 150 & 88.399 \\
Body-Corner Retrieval & Actual body corners & Actual body corners & 126 & 138 & 88.406 \\
w/o Explicit Geometry Encoding & Explicit geometric inputs masked & Side-probe locations & 128 & 140 & 88.488 \\
Fixed-Query Retrieval & Actual body corners & Side probes (fixed after first estimate) & 131 & 143 & 88.426 \\
\textbf{Body--Scene Grounded Planner} & Actual body corners & Side probes (updated across Flow steps) & \textbf{123} & \textbf{136} & \textbf{88.550} \\
\bottomrule
\end{tabularx}
\par\smallskip
{\footnotesize\raggedright\textit{Note:} Fixed-query retrieval holds query locations at those determined by the first clean-trajectory estimate; Flow states continue to evolve.\par}
\par\medskip
\textit{(b) Local visual-output comparison for the Body--Scene Grounded Planner.}\par\smallskip
\begin{tabularx}{\linewidth}{@{}>{\raggedright\arraybackslash}p{.27\linewidth} r *{5}{>{\centering\arraybackslash}X}@{}}
\toprule
Inference configuration & PDMS \ensuremath{\uparrow} & At-fault collisions~\ensuremath{\downarrow} & Geometric contacts~\ensuremath{\downarrow} & DAC failures~\ensuremath{\downarrow} & TTC failures~\ensuremath{\downarrow} & Joint safety failures~\ensuremath{\downarrow} \\
\midrule
Local Visual Output Disabled & 88.451 & 133 & 146 & 499 & 529 & 993 \\
Local Visual Output Enabled & 88.550 & 123 & 136 & 499 & 504 & 965 \\
\bottomrule
\end{tabularx}
\end{table*}

The principal benefit of body-anchored retrieval is reflected in contact safety. When the query references and visual sampling locations are moved from the trajectory centerline to the actual body corners, at-fault collision scenarios decrease from 137 to 126 and geometric contacts from 150 to 138, while PDMS increases slightly. These findings support the safety value of the combined body-anchored configuration, reflecting the joint change in query references and sampling locations. When the side-probe sampling layout is kept fixed and only vehicle dimensions and corner offsets are masked in the query input, at-fault collision scenarios increase from 123 to 128 and geometric contacts from 136 to 140, indicating the supplementary contribution of explicit geometry encoding.

Side-probe sampling further improves planning performance on top of body-corner queries. Keeping the actual body corners as query anchors while moving the visual sampling locations from the corners to the side probes reduces at-fault collision scenarios from 126 to 123 and geometric contacts from 138 to 136, while increasing PDMS from 88.406 to 88.550. This supports the functional distinction between query anchors and visual sampling locations: actual body corners provide geometric references, whereas side-probe locations retrieve local environmental information.

To assess the contribution of cross-step query updates, we train a fixed-query variant from the same Base Planner, retaining body-corner anchors, side-probe sampling, and RC-F. The variant holds retrieval locations fixed after the first clean-trajectory estimate while continuing to update the Flow states. Compared with this variant, dynamic retrieval increases PDMS from 88.426 to 88.550, while reducing at-fault collisions from 131 to 123, geometric contacts from 143 to 136, and joint safety failures from 985 to 965. Mean raw progress and stopping rates remain nearly unchanged. These results support the incremental value of updating spatial retrieval locations using evolving trajectory estimates.

To examine the functional contribution of the local visual pathway, we fix the weights, observations, and initial noise of the Body--Scene Grounded Planner and set the local visual-attention output to zero at every step of the five-step Flow generation process, leaving all other computations unchanged (Table~\ref{tab:IV}(b)).

Disabling the local visual output increases the numbers of at-fault collision scenarios, TTC failure scenarios, and joint safety failure scenarios by 10, 25, and 28, respectively, while leaving the total number of DAC failures unchanged. These results indicate that local visual information contributes to collision and TTC-related trajectory decisions in the trained planner.

Together, these results support the contributions of body-anchored spatial references, side-probe sampling, and cross-step query updates to safe planning, while showing that local visual information influences safety-related trajectory generation.

\subsubsection{RC-F}

The RC-F ablations use the Body--Scene Grounded Planner as a reference to examine the effects of final GT supervision, reference behavior preservation, and the source of query conditioning during training. All configurations share the same initialization, dataset, sampling strategy, and learning-rate schedule. The \emph{w/o Final GT Loss} configuration removes the final GT term from the high-risk objective. The \emph{w/o Rollout Behavior Preservation} configuration removes intermediate-state and final-trajectory preservation while retaining only the local reference velocity-matching objective. \emph{Reference-Conditioned Queries} uses clean-trajectory estimates from preceding Flow steps of the frozen reference policy to construct spatial queries during training, while retaining the trainable planner's five-step generation chain and the complete RC-F objective.

\begin{table}[!tbp]
\caption{Ablation of RC-F.}
\label{tab:V}
\centering
\setlength{\tabcolsep}{3pt}
\renewcommand{\arraystretch}{1.15}
\begin{tabularx}{\linewidth}{@{}>{\raggedright\arraybackslash}X >{\raggedleft\arraybackslash}p{.14\linewidth} >{\raggedleft\arraybackslash}p{.17\linewidth} >{\raggedleft\arraybackslash}p{.18\linewidth}@{}}
\toprule
Training configuration & PDMS \ensuremath{\uparrow} & At-fault collisions~\ensuremath{\downarrow} & Introduced collisions~\ensuremath{\downarrow} \\
\midrule
w/o Final GT Loss & 87.323 & 163 & 3 \\
w/o Rollout Behavior Preservation & 88.750 & 113 & 26 \\
Reference-Conditioned Queries & 88.487 & 129 & 6 \\
Body--Scene Grounded Planner & 88.550 & 123 & 3 \\
\bottomrule
\end{tabularx}
\par\vspace{2pt}
\begin{minipage}{\linewidth}
\footnotesize\raggedright
\emph{Note:} Introduced collisions are measured relative to the Base Planner.
\end{minipage}
\end{table}

In Table~\ref{tab:V}, the Body--Scene Grounded Planner achieves a PDMS of 88.550 with 123 at-fault collision scenarios. Removing the final GT loss yields a PDMS of 87.323 with 163 at-fault collision scenarios. Retaining only the local reference velocity-matching objective results in a higher PDMS of 88.750 and fewer at-fault collision scenarios (113), but introduces 26 at-fault collision scenarios relative to the Base Planner, compared with 3 for the Body--Scene Grounded Planner.

The query-source comparison during training further examines the relationship between spatial retrieval and generated behavior. Reference-Conditioned Queries achieves a PDMS of 88.487 with 129 at-fault collision scenarios, slightly below the Body--Scene Grounded Planner, which constructs queries from its own trajectory estimates. This provides evidence supporting the alignment between query conditions during training and the trajectory generation behavior of the trainable planner.

\begin{figure}[!t]
\centering
\includegraphics[width=\columnwidth]{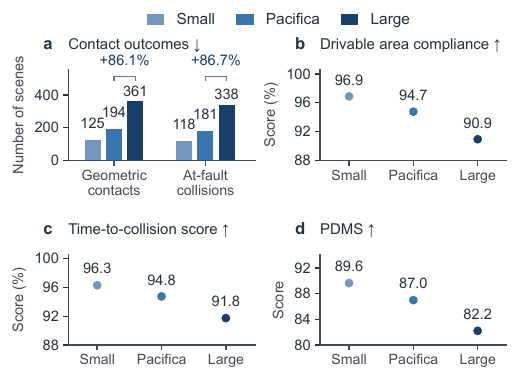}
\caption{Vehicle-geometry counterfactual evaluation on NAVTEST. Fixed base-planner trajectories and simulated states are evaluated under Small, Pacifica, and Large vehicle footprints. (a) Geometric contacts and at-fault collisions; (b) DAC; (c) TTC; and (d) PDMS. Percentage changes in (a) are relative to Pacifica.}
\label{fig:5}
\end{figure}

\begin{figure}[!tbp]
\centering
\includegraphics[width=\columnwidth]{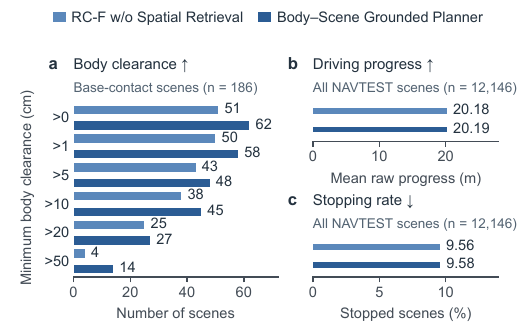}
\caption{Body clearance and driving progress on NAVTEST. (a) Number of Base Planner contact scenes (\(n=186\), across 54 logs) whose minimum body clearance over \(t>0\) exceeds each threshold. (b) Mean raw progress and (c) stopping rate over all 12,146 scenes. A stop is defined by a terminal speed below 0.5 m/s.}
\label{fig:6}
\end{figure}

\begin{figure*}[!t]
\centering
\begin{minipage}[t]{0.49\textwidth}
\centering
\includegraphics[width=\linewidth]{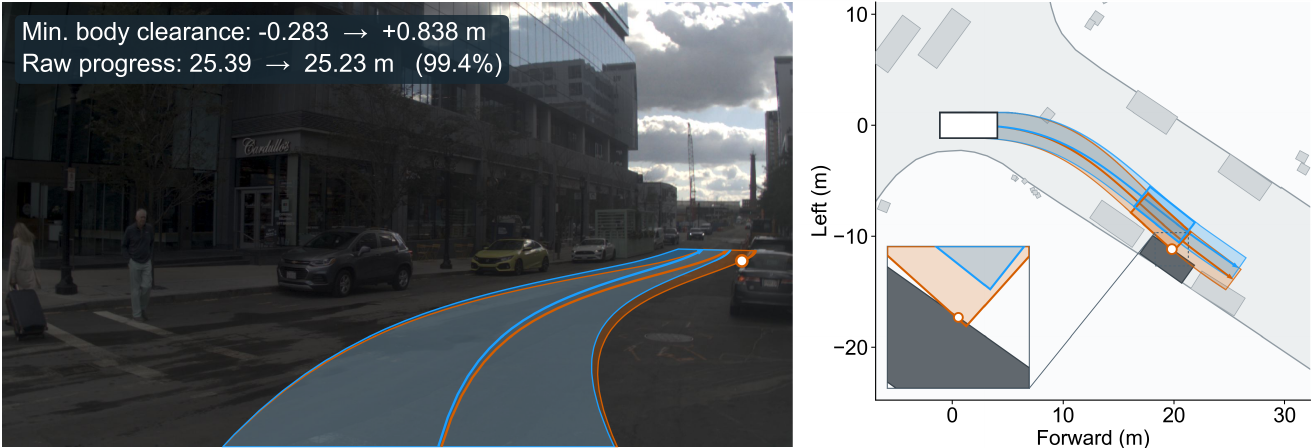}
\par\smallskip (a) Right turn near another vehicle
\end{minipage}\hfill
\begin{minipage}[t]{0.49\textwidth}
\centering
\includegraphics[width=\linewidth]{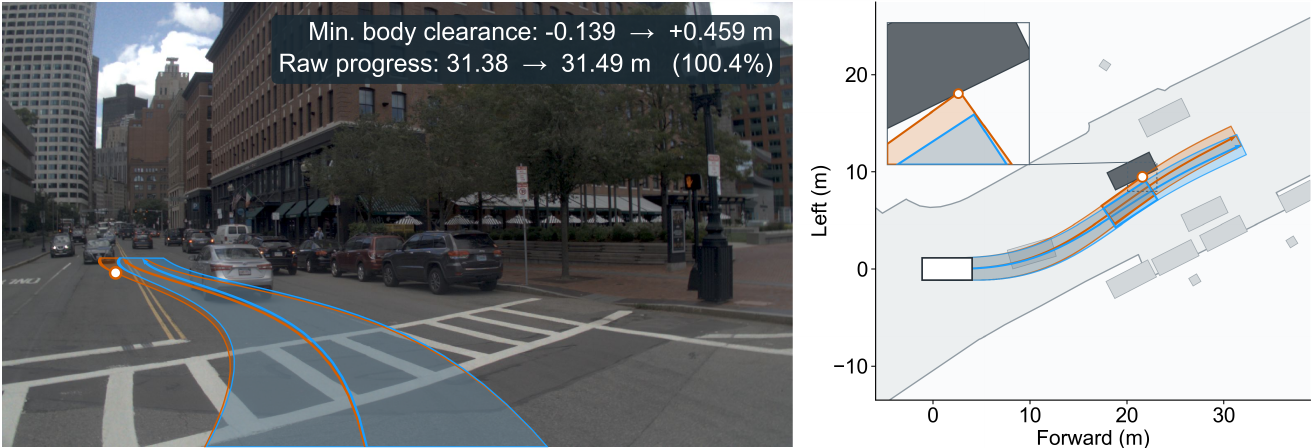}
\par\smallskip (b) Intersection turn
\end{minipage}\par\medskip
\begin{minipage}[t]{0.49\textwidth}
\centering
\includegraphics[width=\linewidth]{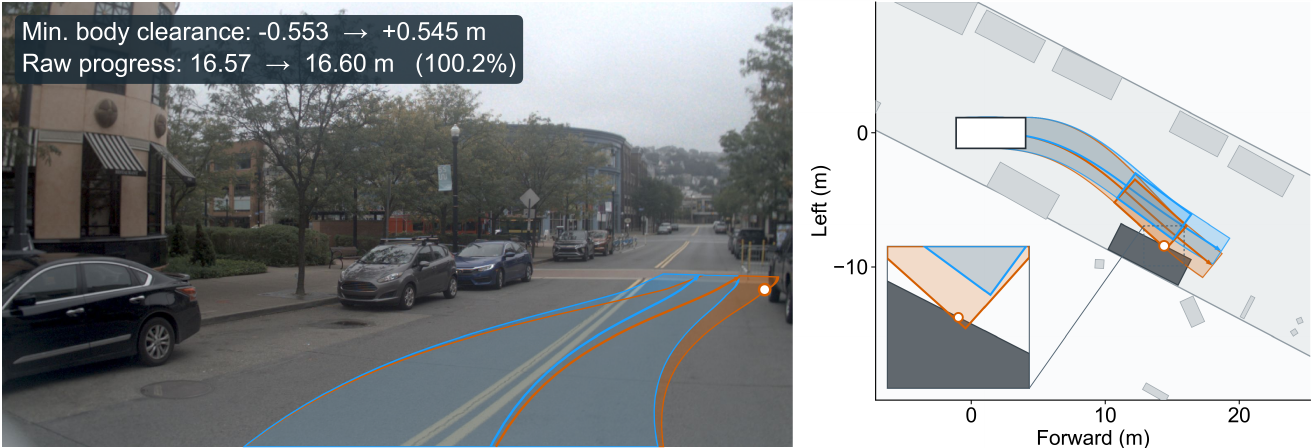}
\par\smallskip (c) Right turn near another vehicle
\end{minipage}\hfill
\begin{minipage}[t]{0.49\textwidth}
\centering
\includegraphics[width=\linewidth]{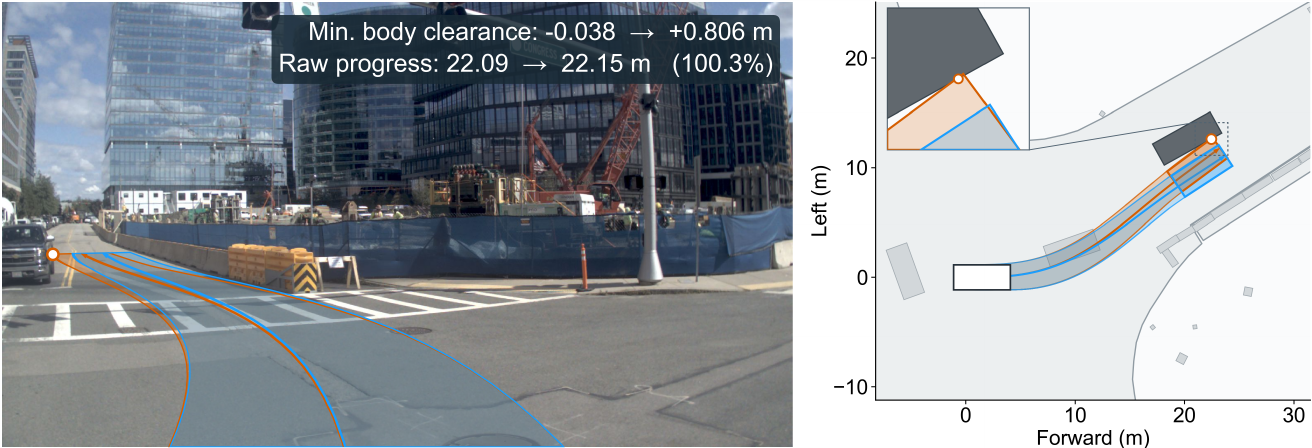}
\par\smallskip (d) Roadworks junction
\end{minipage}\par\medskip
\begin{minipage}[t]{0.49\textwidth}
\centering
\includegraphics[width=\linewidth]{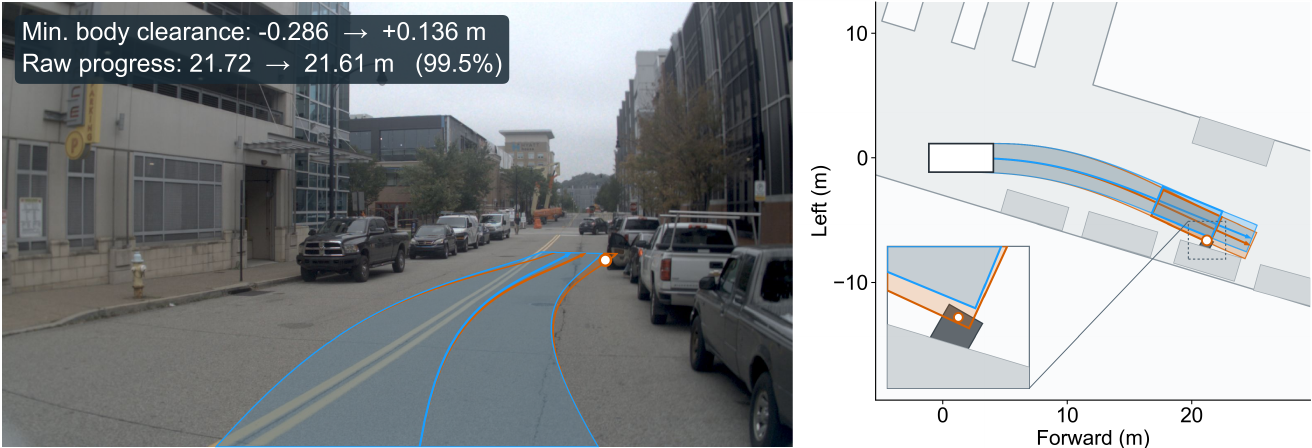}
\par\smallskip (e) Passenger opening a door to exit
\end{minipage}\hfill
\begin{minipage}[t]{0.49\textwidth}
\centering
\includegraphics[width=\linewidth]{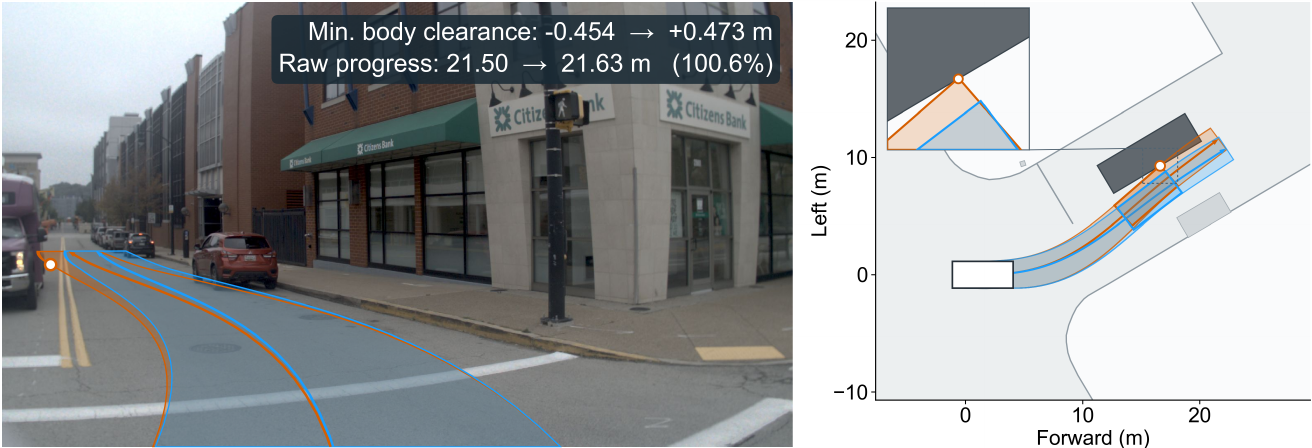}
\par\smallskip (f) Turning near a long vehicle
\end{minipage}\par\medskip
\caption{Case studies of body contact avoidance. Orange and blue denote the Base Planner and Body--Scene Grounded Planner, respectively. Each case shows front-view trajectory and swept-body overlays alongside a BEV comparison at the Base Planner's first contact time, with a local enlargement. In (e), the comparison uses the passenger's annotated footprint. Annotations report minimum signed body clearance and raw progress, with percentages indicating progress retention.}
\label{fig:7}
\end{figure*}

\subsection{Vehicle-Embodied Spatial Feasibility Analysis}
\label{sec:V-D}

\subsubsection{Vehicle-Geometry Counterfactual Evaluation}

Vehicle-geometry counterfactual analysis quantifies the safety differences induced by changes in vehicle-body occupancy for the same planned trajectories. We keep the rear-axle trajectories predicted by the Base Planner, the environment, and the simulated motion states fixed, varying only the planar vehicle footprint used for evaluation. The vehicle geometries \((W,L_f,L_r)\) for Small, Pacifica, and Large are \((2.076,3.524,0.813)\), \((2.297,4.049,1.127)\), and \((2.474,4.773,1.930)\,\mathrm m\), respectively.

Changing vehicle-body geometry alters contact risk even for an unchanged trajectory: the number of geometric contacts under the Large footprint is 86.1\% higher than under Pacifica (Fig.~\ref{fig:5}(a)). This influence also extends to scenarios without direct geometric contact. When changing the vehicle geometry from Pacifica to Large, 462 of the 471 newly introduced DAC failures and 309 of the 366 newly introduced TTC failures occur without geometric contact (Fig.~\ref{fig:5}(b) and (c)). These results indicate that sensitivity to vehicle geometry is reflected not only in direct contacts but also in drivable-area constraints and potential collision risks.

Counterfactual changes in safety status are further used to identify spatially sensitive scenarios (Table~\ref{tab:VI}). Joint safety is defined by \(NC=DAC=TTC=1\), while a strict failure occurs when any of these metrics equals 0. Cases with \(NC=0.5\) and both other metrics equal to 1 are not counted as strict failures. Training scenarios that are jointly safe under Pacifica but exhibit strict failures under Large are combined, after eligibility screening, with low-body-clearance scenarios to construct HardSafe. Detailed criteria are provided in Section~\ref{sec:V-A} and Appendix~\ref{app:B}. Training samples are constructed exclusively from NAVTRAIN, and NAVTEST counterfactual results are used only for analysis.

\begin{table}[!tbp]
\caption{Scenario categories identified by vehicle-geometry counterfactual analysis.}
\label{tab:VI}
\centering
\setlength{\tabcolsep}{3pt}
\renewcommand{\arraystretch}{1.15}
\begin{tabularx}{\linewidth}{@{}>{\raggedright\arraybackslash}Xrr@{}}
\toprule
Scenario category & TRAIN & VAL \\
\midrule
Small safe \ensuremath{\rightarrow} Pacifica fail & 2,226 & 651 \\
Pacifica safe \ensuremath{\rightarrow} Large fail & 5,994 & 1,349 \\
Jointly safe under all three geometries & 74,587 & 14,905 \\
\bottomrule
\end{tabularx}
\end{table}

\subsubsection{Body Clearance and Progress}

This subsection compares RC-F w/o Spatial Retrieval and the Body--Scene Grounded Planner to determine whether the safety gains from spatial retrieval correspond to improvements in physical vehicle-body clearance. Scenarios are grouped according to the frozen Base Planner. Using simulated states sampled at 0.1-s intervals, we compute the minimum signed clearance between the Pacifica body and surrounding obstacles over the entire prediction horizon for \(t>0\). Positive, zero, and negative values indicate separation, contact, and penetration, respectively. After excluding 8 scenarios with initial overlap and 25 scenarios without obstacle geometry, 12,113 scenarios remain for analysis, including 186 Base Planner contact scenarios.

As shown in Fig.~\ref{fig:6}(a), the Body--Scene Grounded Planner increases the number of scenarios maintaining positive clearance throughout the prediction horizon from 51 to 62, compared with RC-F w/o Spatial Retrieval. The number of scenarios with minimum clearance exceeding 10 cm increases from 38 to 45. Spatial retrieval increases the number of scenarios satisfying the clearance requirement at all six evaluated clearance thresholds.

The improvement in body clearance is accompanied by the preservation of overall driving progress (Fig.~\ref{fig:6}(b--c)). RC-F w/o Spatial Retrieval and the Body--Scene Grounded Planner achieve mean raw progress values of 20.18 m and 20.19 m, respectively, and stopping rates of 9.56\% and 9.58\%. These results indicate that, under the same RC-F objective, spatial retrieval improves body clearance without materially reducing overall driving efficiency.

\FloatBarrier
\subsection{Case Studies}
\label{sec:V-E}

Fig.~\ref{fig:7} compares the Base Planner with the Body--Scene Grounded Planner across six cases, illustrating how trajectory adjustments alter the swept vehicle-body occupancy and available clearance.

In the roadworks junction (Fig.~\ref{fig:7}(d)) and the turn near a long vehicle (Fig.~\ref{fig:7}(f)), the two planned trajectories have similar overall shapes, but the Base Planner brings a vehicle-body corner into contact with a nearby vehicle. The Body--Scene Grounded Planner adjusts the turning path to separate the swept body boundary from the obstacle.

The scenario involving a passenger opening a vehicle door to exit (Fig.~\ref{fig:7}(e)) further illustrates locally constrained lateral clearance. The Base Planner's body boundary overlaps the passenger's footprint, whereas the Body--Scene Grounded Planner avoids the contact through a lateral trajectory adjustment. Across all six cases, body contact is eliminated, with raw progress retention ranging from 99.4\% to 100.6\%, indicating that the improved clearance is achieved while maintaining nearly the same driving progress.

\FloatBarrier

\section{Conclusion}
\label{sec:VI}

This paper presented FeaXDrive2, an end-to-end generative driving planning framework designed to enhance vehicle-embodied spatial feasibility. The proposed body-anchored spatial retrieval mechanism uses actual vehicle-body geometry and evolving trajectory estimates to dynamically acquire relevant environmental visual information, allowing body--scene spatial context to guide subsequent Flow generation. RC-F further combines high-risk trajectory supervision with reference behavior preservation, followed by planning-supervised VLM--Action Alignment to jointly adapt multimodal conditioning and action generation. Vehicle-geometry counterfactual evaluation demonstrates that identical rear-axle trajectories can exhibit different contact outcomes and safety-metric outcomes under different body footprints. On NAVSIM v1 NAVTEST, FeaXDrive2 improves PDMS from 87.0 to 88.8 while reducing at-fault collision scenarios from 181 to 115, a 36.5\% reduction. Under the NAVSIM v2 single-stage evaluation on the same scenarios, EPDMS increases from 86.5 to 88.3. Spatial retrieval ablations and body-clearance analysis further support the contribution of body--scene spatial grounding to safer vehicle passage. These findings highlight the importance of coupled vehicle--environment spatial understanding for safer generative driving.

\appendices
\section{Auxiliary Objectives and Trajectory Error Computation}
\label{app:A}

RC-F uses local velocity objectives in addition to the rollout-based losses defined in Section~\ref{sec:IV-C}. A discrete Flow index $i$ is sampled uniformly from $\{0,\ldots,K-1\}$, with $s_i=i/K$. For the initial noise $z_0$ used in the current rollout, an auxiliary interpolation state is constructed as

\begin{equation}
\label{eq:16}
b_i=(1-s_i)z_0+s_i\bar z.
\end{equation}

For high-risk samples, $\bar z=\operatorname{Norm}(\tau^{\mathrm{GT}})$. For behavior preservation samples, $\bar z=\operatorname{Norm}(\tau_{\mathrm{base}}^{\mathrm{fixed}})$, where $\tau_{\mathrm{base}}^{\mathrm{fixed}}$ is a base-planner trajectory precomputed using fixed noise. This precomputed trajectory defines only the endpoint of the auxiliary interpolation path. The reference states and final reference trajectory used for rollout-based preservation are generated separately by the frozen base planner under the current initial noise $z_0$.

The trainable planner evaluates its local velocity at $b_i$, conditioned on the spatial context constructed from the trajectory estimate $r^{(i)}$ produced by its preceding Flow steps. The frozen reference planner evaluates its velocity at the same interpolated state, generation time, and scene condition. Denoting their velocity predictions by $v_{\theta,\phi}^{[i]}(b_i)$ and $v_{\mathrm{ref}}^{[i]}(b_i)$, the auxiliary losses are

\begin{equation}
\label{eq:17}
\begin{aligned}
\ell_{\mathrm{FM}}
&=\frac{1}{3H}\left\|
 v_{\theta,\phi}^{[i]}(b_i)
 -\bigl(\operatorname{Norm}(\tau^{\mathrm{GT}})-z_0\bigr)
\right\|_F^2,\\
\ell_{\mathrm{velocity}}
&=\frac{1}{3H}\left\|
 v_{\theta,\phi}^{[i]}(b_i)-v_{\mathrm{ref}}^{[i]}(b_i)
\right\|_F^2.
\end{aligned}
\end{equation}

The first loss applies to high-risk samples, whereas the second applies to behavior preservation samples. Both are evaluated at constructed interpolation states, rather than replacing the intermediate-state and final-trajectory losses computed on the actual five-step rollout.

For a trajectory error matrix $E\in\mathbb R^{H\times3}$ in physical units, the coordinate-scaled Smooth L1 loss is

\begin{equation}
\label{eq:18}
\rho(E)=\frac{1}{3H}\sum_{h=1}^{H}\sum_{\chi\in\{x,y,\psi\}}
\operatorname{SmoothL1}_{\beta=1}\!\left(\frac{E_{h,\chi}}{\sigma_\chi}\right),
\end{equation}

where $(\sigma_x,\sigma_y,\sigma_\psi)=(1\,\mathrm m,0.1\,\mathrm m,0.02\,\mathrm{rad})$. Differences between normalized states are converted to physical units by right-multiplication with the scaling matrix $S$; their common offset cancels. Intermediate-state losses are computed without clipping or heading-angle wrapping. For final-trajectory errors $e(\tau,\tau')$, position components are subtracted directly and heading differences are wrapped to $[-\pi,\pi]$.

With five Euler steps, the intermediate clean-trajectory weights in Section~\ref{sec:IV-C} are $(w_0,w_1,w_2,w_3)=(0.1,\allowbreak 0.2,\allowbreak 0.3,\allowbreak 0.4)$.

\section{Training Sample Construction}
\label{app:B}

The NAVTRAIN split contains 85,109 original scenes. The four-role safety-adaptation pool contains 82,678 scenes after excluding 2,431 scenes: 29 with invalid initial states, 20 with geometric contact but without a qualifying collision role, 2,357 with other safety failures that do not meet the role criteria, and 13 collision and 12 near-collision candidates whose demonstrations fail the supervision filter. Ineligible high-risk candidates are not reassigned to the behavior preservation pool.

Collision candidates satisfy $\mathrm{NC}<1$ and exhibit geometric contact. Near-collision candidates satisfy $\mathrm{NC}=1$ and $\mathrm{TTC}=0$ without geometric contact; $\mathrm{DAC}=1$ is not required for the predicted trajectory. Demonstrations used for high-risk supervision must satisfy $\mathrm{NC}=\mathrm{DAC}=\mathrm{TTC}=1$, exhibit no geometric contact under Pacifica geometry, and retain at least 50\% of the base planner's normalized and raw progress. Demonstrations are additionally rejected when a larger tested footprint is jointly safe while a smaller footprint fails the joint-safety criterion. This consistency check applies to demonstration eligibility, not to the counterfactual selection of hard-safe scenes.

HardSafe is formed from base-safe scenes without geometric contact that satisfy either of two conditions: a minimum signed body clearance in $(0,0.5]$ m, or a strict safety failure under the Large footprint despite joint safety under Pacifica. For training sample construction, clearance is computed at 0.1-s intervals over 0--4 s; scenes with inconsistent contact and zero-clearance flags are excluded. The strict counterfactual failure rule uses $\mathrm{NC}$, $\mathrm{DAC}$, and $\mathrm{TTC}$ as defined in Section~\ref{sec:V-D}.

Of the 5,994 NAVTRAIN candidates satisfying the Pacifica-safe-to-Large-failure transition in Table~\ref{tab:VI}, 5,991 pass the final role-eligibility checks. The low-clearance criterion identifies 3,596 eligible scenes, with 914 scenes satisfying both criteria. The resulting HardSafe pool contains $5{,}991+3{,}596-914=8{,}673$ scenes. The remaining eligible base-safe scenes form the Normal pool. These training selections use NAVTRAIN only; NAVTEST counterfactual results are used for analysis and do not contribute training samples.

For VLM--Action Alignment, training candidates are reassessed using the frozen Body--Scene Grounded Planner. Failed scenes enter high-risk supervision only when their demonstrations remain safety-qualified and satisfy the progress criteria relative to this planner. Safe scenes retain existing hard-safe or footprint-sensitive labels; a positive body clearance of at most 0.5 m also assigns the hard-safe role. The remaining eligible safe scenes form the normal group. High-risk, hard-safe, and normal samples are drawn in proportions of 10\%, 25\%, and 65\%, respectively.

\section{Architecture and Sampling Details}
\label{app:C}

Action Flow-DiT has 16 blocks, with scene cross-attention at zero-based indices $[1,3,5,7,9,11,13,15]$ and spatial context injection at $[9,11,13,15]$. RC-F updates spatial retrieval, self-attention in blocks 12 and 14, and feedforward and adaptive layer-normalization components in blocks 12--15. Scene cross-attention, the VLM, the condition adapter, and all remaining parameters are frozen.

VLM--Action Alignment updates the same Action Flow-DiT components and q/v-projection LoRA parameters in the last eight language-model layers, with rank $r=16$ and scaling $\alpha=32$. Spatial retrieval, the visual encoder, the condition adapter, and all remaining parameters are frozen. Training holds reasoning-token sequences from the unadapted VLM fixed and recomputes their hidden representations differentiably. At inference, the adapted VLM generates reasoning sequences autoregressively to produce conditioning representations. The VLM uses BF16; LoRA parameters, optimizer states, and the planner use FP32.

Each nine-dimensional body-corner input uses a two-layer SiLU MLP and a corner-type embedding. Current-frame features from three front-view cameras are projected from 1,152 channels to $D=384$. Images are resized from $1920\times1080$ to $576\times320$ pixels (width $\times$ height), with features resampled to a $16\times16$ grid per camera. Each valid side-probe projection supplies a $3\times3$ bilinear sampling neighborhood at grid offsets $\{-1,0,1\}\times\{-1,0,1\}$. Validity requires positive camera depth, valid lens-distortion mapping, and in-image coordinates. Out-of-grid neighbors use border padding; invalid projections are masked, with a learnable null-visual token for queries lacking valid samples. The side-probe offset is $d=1.5\,\mathrm m$.

\emph{Local visual-output intervention.} For Table~\ref{tab:IV}(b), the local visual-attention module's projected output is zeroed at every Flow step before residual addition to each body-corner query. All other operations remain unchanged.

Trajectory normalization is applied independently to each future pose:

\begin{equation}
\label{eq:19}
\begin{aligned}
\operatorname{Norm}(\tau)&=\bigl(\tau-\mathbf 1_H\mu^\top\bigr)S^{-1},\\
\operatorname{Denorm}(z)&=zS+\mathbf 1_H\mu^\top,
\end{aligned}
\end{equation}

where $S=\operatorname{diag}(33.37,21,1.765)$ and $\mu=(31.8,\allowbreak 1.32,\allowbreak 0.095)^\top$. Clipping to $[-1,1]$ is applied before denormalization at the final trajectory output and when constructing the next spatial-query condition.

Base planner learning uses a global batch size of 128, learning rate $10^{-4}$, and BF16 precision.

RC-F uses two NVIDIA L40 GPUs with 16 samples per GPU (global batch 32). Its 4,000-step cosine schedule includes 200 warm-up steps and decays to one-tenth of the peak. Peak rates are $10^{-4}$ for spatial retrieval and $10^{-5}$ for the selected backbone parameters. AdamW \cite{loshchilov2019adamw} uses momentum coefficients $(0.9,0.95)$, weight decay $10^{-4}$, and gradient clipping at 1.0. RC-F training and planner inference use FP32 with random seed 0.

RC-F shuffles and cycles through each role independently. In the role order of Table~\ref{tab:I}, each five-update cycle uses quotas $(8,5,8,11)$ for four updates and $(8,4,8,12)$ for one. Losses are averaged over the global batch.

VLM--Action Alignment keeps the high-risk loss unchanged and scales all behavior-preservation loss weights by 0.2. The global batch size is 16. Peak learning rates are $3\times10^{-6}$ for language-model LoRA and $2\times10^{-6}$ for the selected Action Flow-DiT parameters, with warm-up and cosine decay to one-tenth of their peaks.

\bibliographystyle{IEEEtran-feaxdrive}
\bibliography{references}
\end{document}